\documentclass[journal=jacsat,manuscript=article]{achemso}

\usepackage{braket}
\usepackage{booktabs}
\usepackage{amsfonts}
\usepackage{hyperref}
\usepackage{lineno}
\usepackage{tabularx}
\usepackage{graphicx}

\usepackage[version=3]{mhchem} 

\usepackage{amsmath}

\author{Alessio Fallani}
\affiliation[ULUX]{Department of Physics and Materials Science, University of Luxembourg, L-1511 Luxembourg City, Luxembourg}
\alsoaffiliation[Janssen]{Drug Discovery Data Sciences, Janssen Pharmaceutica NV, Turnhoutseweg 30, 2340 Beerse, Belgium}
\author{Ramil Nugmanov}
\affiliation[Janssen]{Drug Discovery Data Sciences, Janssen Pharmaceutica NV, Turnhoutseweg 30, 2340 Beerse, Belgium}
\email{rnugmano@its.jnj.com}
\author{Jose Arjona-Medina}
\affiliation[Janssen]{Drug Discovery Data Sciences, Janssen Pharmaceutica NV, Turnhoutseweg 30, 2340 Beerse, Belgium}
\author{J\"org Kurt Wegner}
\affiliation[JnJ]{Johnson \& Johnson Innovative Medicine, 301 Binney Street, Cambridge, MA 02142, USA}
\author{Alexandre Tkatchenko}
\affiliation[ULUX]{Department of Physics and Materials Science, University of Luxembourg, L-1511 Luxembourg City, Luxembourg}
\author{Kostiantyn Chernichenko}
\affiliation[Janssen]{Drug Discovery Data Sciences, Janssen Pharmaceutica NV, Turnhoutseweg 30, 2340 Beerse, Belgium}
\email{kcherni1@its.jnj.com}
\title[An \textsf{achemso} demo]
  {Pretraining Graph Transformers with Atom-in-a-Molecule Quantum Properties for Improved ADMET Modeling}

\abbreviations{IR,NMR,UV}

\begin{document}


\begin{abstract}

We evaluate the impact of pretraining Graph Transformer architectures on atom-level quantum-mechanical features for the modeling of absorption, distribution, metabolism, excretion, and toxicity (ADMET) properties of drug-like compounds.
We compare this pretraining strategy with two others: one based on molecular quantum properties (specifically the HOMO-LUMO gap) and one using a self-supervised atom masking technique. 
After fine-tuning on Therapeutic Data Commons ADMET datasets, we evaluate the performance improvement in the different models observing that models pretrained with atomic quantum mechanical properties produce in general better results. We then analyse the latent representations and observe that the supervised strategies preserve the pretraining information after finetuning and that different pretrainings produce different trends in latent expressivity across layers. Furthermore, we find that models pretrained on atomic quantum mechanical properties capture more low-frequency laplacian eigenmodes of the input graph via the attention weights and produce better representations of atomic environments within the molecule. 
Application of the analysis to a much larger non-public dataset for microsomal clearance illustrates generalizability of the studied indicators. In this case the performances of the models are in accordance with the representation analysis and highlight, especially for the case of masking pretraining and atom-level quantum property pretraining, how model types with similar performance on public benchmarks can have different performances on large scale pharmaceutical data.

\end{abstract}

\section{Introduction}
Effectively representing molecules for modeling applications is a fundamental challenge in cheminformatics and machine learning, which lead to the development of various representation methods. In the realm of precomputed representations, different approaches are utilized depending on the available input data. Fingerprints, which encode the presence or absence of substructures and certain chemical properties in binary vectors, are commonly used in cheminformatics, particularly when 3D data is not available. On the other hand, physics-inspired representations like the Coulomb Matrix \cite{coulombmatrix}, Bag of Bonds\cite{BoB}, SLATM \cite{SLATM} and many others are more frequently employed for representing 3D geometries and physical properties. 
With the advent of deep learning, the potential to learn representations directly from data has become increasingly apparent\cite{Chen2018, Jayatunga2022, Bule2021, Li2022, Chuang2020, Born2023}. These learned representations, shaped by the training of deep neural networks, enable the transformation of input data into a latent space where relevant features can be distilled in different ways for specific tasks\cite{Wang2022, Kaufman2024, encoderfingerprints}. For instance, contrastive learning techniques have been employed to learn representations including information from other modalities such as image information\cite{contrastiveimage} and knowledge-graph information\cite{KGcontrastive}. Similarly, other learning approaches have been developed to integrate multimodal data with chemical structures, such as medical records \cite{medicalrecords}, natural language \cite{nlpmultimodal} or pooling together sequence graph and geometry information \cite{seqgraphgeom}. Moreover, this approach often allows for the incorporation of invariance or equivariance with respect to particular transformations, enhancing the robustness and accuracy of the models \cite{torchani, schnet, nequip}. 
However, despite the remarkable successes achieved, these methodologies still present some limitations that need to be carefully evaluated \cite{Deng:23}. In particular, challenges such as data scarcity\cite{Dourew2023} and generalizability remain pertinent concerns in the field \cite{Born2023, Glavatskikh2019, Ektefaie2024, Broccatelli2022, Huang2021, Valentinrew}. To address these challenges, the concept of pretraining models on related tasks or employing self-supervised learning strategies has gained significant traction. The success of this methodology is evident, for example, in the realm of natural language processing, where overparameterized large language models (LLMs) are pretrained on a wide corpus of data, and then made available for fine-tuning with minimal resources and small datasets on specific tasks \cite{chatgpt4, llama}. Following a similar paradigm, in the context of molecular representation learning this technique has been explored as a mean to enhance model generalizability and performance across various downstream tasks \cite{Kaufman2024, Wang2023, Xia2023iclr, Xia2023ijcai, Hu2020Strategies}. The selection of pretraining data and tasks, though, is not trivial. The data should be such that: (i) it is available or can be generated at scale and (ii) provide fundamental information about molecular properties and behaviour. A natural choice following these criteria is quantum mechanical (QM) reference data, as it is known to be related to fundamental aspects of molecular behavior \cite{Beck2005, quantumtox, fukuibook, atomicdescriptordesign} with profound implications in biochemical research and it only requires computational resources for production at scale, being in fact already present in an increasing number of public datasets \cite{Hoja2021, MedranoSandonas2024, Isert2022, OQMD, NakataPM6, OpenCatalyst, sgdml, HarvardCleanEnergy, FreeSolv, BindingDB, Ani1x, Ani1}. Studies utilizing both atomic and molecular QM properties for pretraining have been already carried out in multiple excellent works \cite{minimol, Kim2024, raja2024on} using different molecular representations and architectures. In these studies some degree of improvement on various downstream tasks is shown, but the conclusions are often based solely on the modelling benchmark data and explained by knowledge transfer between tasks, limiting the understanding of the real impact that different pretraining methods can have on the representations learned by the models. In this context, our study aims to bridge this gap and proposes a series of studies focused on investigating the impact that pretraining on atom-level quantum-mechanical (QM) properties has on the representation learned by a Graphormer neural network \cite{Ying:21} when compared to other commonly employed pretraining strategies. The evaluation is carried out on public benchmark data and multiple analysis are performed on fine-tuned models with the addition of a test on internal microsomal clearance data. The results show that models pretrained atom-level QM properties result in better representatoins under multiple indicators, and that the ranking based on those indicators matches better with the results on the larger internal dataset rather than with the results on the benchmark.

\section{Methods}\label{methods}
We consider a custom implementation of Graphormer \cite{Nugmanov2022, Ying:21} as an instance of network that belongs to the increasingly popular family of Graph Transformers (GTs) \cite{muller2024attending}, models that generally utilize a transformer architecture on 2D graph input data.  
As a comparison for the models pretrained on atomic QM properties, besides training the model without pretraining, we consider masking pretraining (atom-level self-supervised method) using the same dataset employed for the atom-resolved QM properties \cite{Guan2021}, and pretraining on on a much bigger dataset of a molecular property, HOMO-LUMO gap (HLG) calculated by QM methods \cite{Nakata2017}. This choice is dictated by the approximate matching of the overall number of atomic properties (molecules times non-hydrogen atoms) with the number of datapoints for the HLG (one per molecule). The pretrainings were followed by fine-tuning on individual target downstream tasks from  absorption, distribution, metabolism, excretion, and toxicity (ADMET) benchmark datasets of the Therapeutics Data Commons (TDC) \cite{Huang2022}. They represent the key properties relevant to pharmacokinetics and pharmacodynamics of drugs. Other than comparing the results of these different pretraining strategies on the benchmark metrics, the studies carried out in this work investigate multiple aspects of learned latent representations. Namely, we evaluate the conservation of the pretraining information after fine-tuning, analyse expressivity of the latent representation across layers and sensitivity of the receptive field of the obtained atomic representations. We also propose and perform a novel spectral analysis of the Attention Rollout matrix \cite{rollout}, that studies its relation to the graph-Laplacian eigenmodes of the input molecule. Furthermore, the pretraining methodologies were utilized in the modeling of an internal company dataset of microsomal clearance (which contains much more data than its public TDC counterpart) that revealed the limitations of using only public benchmark metrics for methodology evaluation.
In this section we will describe in detail the model, the datasets, the methods used for pretraining and finetuning as well as each of the analyses done on the fine-tuned models.

\paragraph{\textbf{Model description}} Graphormer is a GT where the input molecule is seen as a graph with atoms as nodes and bonds as edges. This model in general works by encoding the atoms in the molecule as tokens based on their atom type, and then repeatedly applying self-attention layers with an internal bias term before the softmax function. This term is based on the topological distance matrix of the molecular graph and allows to encode the structural information of the molecular graph.
In particular, the network employed in this work is an implementation of Graphormer from \cite{Nugmanov2022}, inspired by the work \cite{Ying:21}. In this implementation the centrality encoder is adapted from using only explicit neighbours to including both explicit atoms and implicit hydrogens. As a result of the combination of this modified centrality encoding together with the usual atom type encoder, the hybridization of atoms is handled implicitly. For this reason this implementation does not present any edge encoder component.
For what concerns the choice of hyperparameters, we did not run hyperparameter tuning experiments as absolute performance is not the focus of this work. We purposely chose 20 hidden layers, a higher number than usually found in similar architectures, while maintaining a number of parameters that is comparable with other Transformer-based implementations previously introduced\cite{Ying:21, Fabian2020} ($\sim$ 10M parameters). This choice is done in order to study the effects of the pretraining strategies on very deep models, considering quantities closely related to known depth-related phenomenons in machine learning literature \cite{muller2024attending, attentionrank, oversquashing_bronstein}, while maintaining reasonable training times and a number of parameters comparable to other models. The rest of the hyperparameters were chosen based on our experience and maintaining the same conditions across all models in both pretraining and fine-tuning stages for fairness of comparison.
Finally, differently from our preliminary results in \cite{joseramil, icann} we do not employ task specific virtual nodes but rather rely solely on the original implementation in \cite{Nugmanov2022}. This last choice is made to exclude any effect that this technique may have on the final results, especially considering that other GT architectures generally do not employ it. More information on the Graphormer implementation is reported in the SI.

\paragraph{\textbf{Pretraining datasets and methodology}} For pretraining, we used a publicly available dataset \cite{Guan2021} consisting of $\sim$136k organic molecules for a total of over 2M heavy atoms. 
Each molecule is represented by a single conformer initially generated using the Merck Molecular Force Field (MMFF94s) in RDKit library. The geometry for the lowest-lying conformer was then optimized at the GFN2-xtb level of theory followed by refinement of the electronic structure with DFT  (B3LYP/def2svp). The dataset reports several atomic properties: a charge, electrophilic and nucleophilic Fukui indexes, an NMR shielding constant.
Another pretraining dataset, PCQM4Mv2, consists of a single molecular property per molecule, an HLG that was also calculated using quantum chemistry methods \url{https://ogb.stanford.edu/docs/lsc/pcqm4mv2/}. The dataset contains over 2M of molecules and was curated under the PubChemQC project \cite{Nakata2017}. It is important to specify that albeit both datasets also contain the 3D molecular geometries, we only employ the 2D graph chemical structures.

The pretraining on atom-level QM properties is achieved via a regression task by applying a linear layer to the obtained node representations, each corresponding to a heavy (non-hydrogen) atom.
The model is trained on each one of the available atomic properties separately, as well as on all of them at the same time in a multi-task setting. As a result, we obtain from these pretraining efforts $5$ different models.

Pretraining on molecular quantum properties is achieved via a regression task on the values of HLG from the PCQM4Mv2 dataset and where the output is obtained by applying a linear layer to the class token representation at the last layer of the network.

Masking pretraining, instead, is carried out in a similar way to what is usually done in BERT-based models \cite{Fabian:20, BERT}. This procedure entails randomly masking $15$\% of the input graph node tokens by replacing them with the mask token, and then training the model to restore the correct node type from the masked embedding as a multi-class classification task.
This last pretraining is carried out on the molecular structures present in the dataset used for atomic QM properties.

\paragraph{\textbf{Downstream Tasks}}
For the benchmarking of the obtained pretrained models, we used the absorption, distribution, metabolism, excretion, and toxicity (ADMET) group of the TDC dataset, consisting of 9 regression and 13 binary classification tasks for modeling biochemical molecular properties \url{https://tdcommons.ai/benchmark/admet_group/overview/}. The training and testing on this dataset is carried out in the same way as any molecular property modeling. For splittings and evaluation metrics we follow the guidelines of the benchmark group that we consider, hence we refer to \cite{Huang2022}.
All the downstream task trainings followed the same procedure with weights taken from the respective preartained models (without freezing any layer) or randomly initialized for training the scratch model. No multitask training is adopted here and a different model is obtained for each split of each downstream task.
For each combination of downstream task and pretraining, we obtained $5$ models, corresponding to training/validation splits as provided in the benchmark, and reported the final performance as mean and standard deviation over this set. 
In summary, the non-pretrained Graphormer version used as a baseline model was compared with $7$ different pretrained models: one per each of the $4$ atom-resolved QM properties (atomic charges, NMR shielding constants, electrophilic and nucleophilic Fukui function indexes), one pretrained on all atomic properties in a multi-task setting, one for the molecule-level property (HLG), and one for masking node pretraining. Each of the 8 models is then fine-tuned on the 22 tasks from the TDC ADMET benchmark totalling 880 final models with the default 5 train/test splits per each task. 

Finally, we consider a much larger set of proprietary JNJ data containing values of human liver microsome (HLM) intrinsic clearance of $\sim 130k$ compounds measured in two different assays. Fine-tuning on this dataset was conducted in the same way as in the case of the regression tasks in TDC with the difference that two values of clearance were modelled using a multi-task approach. A test set accounts for 20\% of a total dataset size and is obtained as a scaffold split that maximizes Tanimoto distance between the train and test splits. The results are reported both in terms of $R^2$ coefficient and Spearman's correlation coefficient as mean and standard deviation over 3 seeds. 
 
\paragraph{\textbf{Conservation of pretraining information after fine-tuning}}
In order to understand if the fine-tuned models preserve some of the information learned during the supervised pretraining stages or if that amounts only to a different network initialization, we analyse the latent representation obtained in the last layer. In particular, for each fine-tuning task and for each set of differently pretrained models, we freeze the model obtained from one of the seeds and encode a sample of $5000$ molecules from each of the two pretraining datasets. The latent representations are split into equal size train/test sets and fit with the regularized linear regressor from \cite{elasticnet} to reveal to which extent the representation still preserve linear correlation with the pretraining labels. The results are reported in terms of $R^2$ coefficient over the test set averaged across the 22 fine-tuned models. We perform this analysis in an all-to-all fashion: namely for every model against every pretraining task and also considering the models trained from scratch as some correlation may arise from learning representations during the downstream tasks.

\paragraph{Latent expressivity across layers}
The internal representation at each layer of the models is also studied by analysing a quantity introduced in \cite{attentionrank}, which is related to representation rank, that measures how similar are latent token representations. If $\mathrm{GT}_L(\textbf{X})\in\mathbb{R}^{n\times d}$ is the latent representation of an encoded input $\textbf{X}\in\mathbb{R}^{n\times d}$ at layer $L$ of a GT network, this is defined as:
\begin{equation}
    \rho_L = \frac{||\mathrm{res}(\mathrm{GT}_L(\textbf{X}))||_{1, \infty}}{||\mathrm{GT}_L(\textbf{X})||_{1, \infty}}
\end{equation}
with $||(\cdot)||_{1, \infty} = \sqrt{||(\cdot)||_1||(\cdot)||_{\infty}}$, where $\mathrm{res}(\textbf{X}) = \textbf{X} - \textbf{1}\textbf{x}^T$, with $\textbf{x} = \underset{\textbf{x}}{\mathrm{argmin}} ||\textbf{X} - \textbf{1}\textbf{x}^T||$ where $\textbf{x}\in\mathbb{R}^{d}$ and $\textbf{1}\in\mathbb{R}^{n}$.
Namely, this metric measures how close is the representation to the closest representation in $||(\cdot)||$ norm where all $n$ latent representations are equal to the same vector $\textbf{x}$.
We report the value of this quantity across all layers for every model, computed using a random sample of $100$ molecules from each test set of the ADMET tasks.

\begin{figure}
    \centering
    \includegraphics[width=\textwidth]{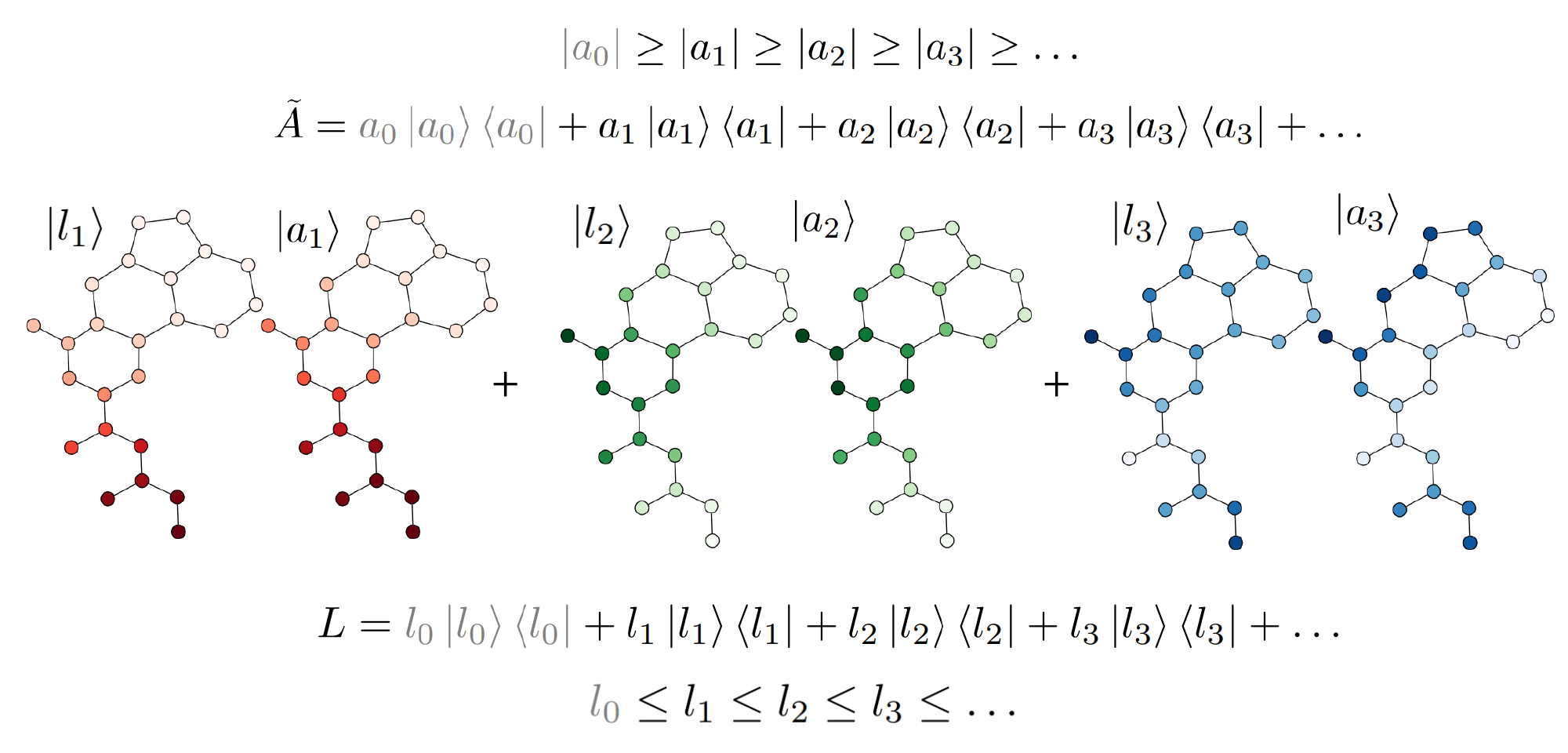}
    \caption{Visual representation for a molecule in the TDC dataset of the comparison between the most relevant eigenvectors of the attention rollout matrix from a model pretrained on atom-level QM properties and the low-frequency eigenvectors of the graph Laplacian associated to the molecular structure. $\ket{a_i}$ are the eigenvectors of the Attention Rollout matrix $\Tilde{A}$ with eigenvalue $a_i$ and $\ket{l_i}$ are the eigenvectors of the graph Laplacian $L$ with eigenvalue $l_i$.} 
    \label{fig: molecule_splitting}
\end{figure}

\paragraph{\textbf{Spectral Analysis of Attention Rollout}}
To have a better understanding of the mechanism behind the pretrained models' improvements, we shift our focus on the analysis of attention weights. 
We aim to understand directions along which an input molecular representation is decomposed when passed through a given model. In order to do so, we start by considering the Attention Rollout matrix \cite{rollout} $\Tilde{A}$ as a proxy for the model's action on the input (see SI for a more detailed motivation). While this is a strong approximation, it provides a number of non-trivial insights (vide infra).
We operate an eigendecomposition of $\Tilde{A}$ (from here on we will make use of the bra-ket notation):
\begin{equation}
\Tilde{A} = \sum_{i=0}^{N-1} a_i\ket{a_i}\bra{a_i} 
\end{equation}
with $a_i\in\mathbb{C}$ and $|a_0|\geq |a_1|\geq ...\geq |a_{N-1}|$ and, based on an empirical observation on one of the pretrained Graphormers (see Fig. \ref{fig: molecule_splitting}), we analyse the similarity of the eigenvectors $\ket{a_i}\in\mathbb{C}^n$ with the eigenvectors of the Laplacian matrix $L$ of the input molecular graph decomposed as 
\begin{equation}
    L = \sum_{i=0}^{N-1} l_i \ket{l_i}\bra{l_i}
\end{equation} 
with $l_i\in\mathbb{R}$, $\ket{l_i}\in\mathbb{R}^n$ and $l_0\leq l_1\leq ...\leq l_{N-1}$. In particular, by considering the overlap matrix $C_{ij} = |\braket{l_i|a_j}|$ we study both how many Laplacian modes are used as models' eigendirections as well as how relevant they are as fraction of the non-trivial spectrum of $\Tilde{A}$ (by non-trivial we mean $i\neq 0$ as by construction $|\braket{l_0|a_0}|=1$ for reasons reported in the SI). This fraction is quantified by $\eta = \frac{\sum_{i\in\mathcal{U}\setminus 0} |a_i|}{\sum_{i=1}^{i=N-1} |a_i|}$ where $\mathcal{U} = \{j | \max_j C_{ij}\geq0.9\; \mathrm{for}\; i \in \left(0,1,2,...,N-1\right)\}$ with $0.9$ being a chosen arbitrary threshold for similarity. Based on these quantities, we define a metric that factors everything together as:
\begin{equation}\label{zeta_metric}
\zeta = \eta \sum_{i=1}^{N-1}\Theta\left(\max_{j}C_{i,j}-0.9\right)
\end{equation}
where $\Theta$ is the Heaviside function. We then evaluate $\zeta$ averaged over the test set of each downstream task reporting per each group of models the distribution across tasks for fixed pretraining condition. This quantity, as noted in the SI, can be loosely interpreted as an indicator of oversmoothing of the input graph information. A higher $\zeta$, in fact, can be intepreted as a higher bandwith in the Fourier space defined by the input graph.

\paragraph{\textbf{Neighbour sensitivity analysis}}
Following the hypothesis that atomic QM properties provide a good description of the atomic environment around each atom, we carry out a sensitivity analysis to understand in every model how much each atomic representation in the last layer is influenced by changes in the input embedding of the $k^{th}$ neighbours within the same molecule. In order to do so, we compute the following quantity for 50 randomly selected molecules from the TDC test sets:
\begin{equation}
    \mathcal{S}_k = \bigg\langle\bigg\langle\sqrt{\sum_{\nu = 0}^{d}\sum_{\mu = 0}^{d}\left(\frac{\delta\mathrm{GT}(\textbf{X})_{i,\nu}}{\delta \textbf{X}_{k,\mu}}\right)^2}\bigg\rangle_{k\in\mathcal{K}_i}\bigg\rangle_{i\in \mathcal{M}},
\end{equation}
where $\mathcal{K}_i$ is the set of $k^{th}$ neighbours of the atom $i$ and $\mathcal{M}$ is the set of the atoms in the molecule.
A closely related quantity has been studied as indicator of oversquashing in graph message-passing neural networks, a phenomenon where bottlenecks in the message passing mechanism prevent proper information propagation \cite{oversquashing_bronstein}. In our case, it is used as measure of receptive field for the representation of atomic environments obtained with the models, and compared across pretraining methodologies. For each molecule the vector of sensitivities $\left[\mathcal{S}_0,\mathcal{S}_1, \dots \right]$ is then standardized subtracting the minimum value and then dividing by the maximum value (which is usually $\mathcal{S}_0$). These vectors are collected for the sampled structures and the selected models, and the behaviour is then analysed from the first to fifth neighbour atom node.

\section{Results}

\paragraph{\textbf{Benchmark results}} Model performances obtained for the downstream tasks are summarized in table \ref{results} where the best model in the tested group is highlighted in bold. We evaluate the best results based on their mean values and then perform a paired t-test to test the hypothesis that they are significantly better than the others. The models for which the null hypothesis cannot be refuted were highlighted with the only exception being the exclusion of two cases where the standard deviation is one order of magnitude higher than for all the other results. 
Overall, while in most cases all the pretraining strategies provide some improvement, pretraining on HLG stands out only for one property, albeit still being among the best models in the group for four more cases. While masking pretraining also significantly outperforms other models only in one case, we find it sharing top performance with other models for ten more downstream tasks. When the models pretrained with atom-level QM properties are considered as a group, we find it to contain the best model overall (at least one better than both masking and HLG) in ten cases, and tying for best model in twenty cases out of twenty-two. Within the group one can see that models pretrained on charges, NMR shifts and all atomic QM properties provide overall a greater number of best results than models pretrained on Fukui functions. Finally, we notice that for the case of solubility, lipophilicity and acute toxicity (LD50) we obtain superior results than the respective best models in the TDC leaderboard.

\begin{table} 
\caption{Global results obtained from the ADMET group of TDC are presented. Each row corresponds to a specific task, along with the metric used for evaluation. Columns represent different pretrainings considered. Highlighted values denote the best performance achieved among our models, based on the average value and t-tests paired across seeds. Additionally, cases where our results surpass in mean value the top-performing model in the TDC leaderboard are marked with an asterisk ($^*$).}
\label{results}
\resizebox{\textwidth}{!}{%
    \begin{tabular}{llcccccccc}
        \toprule
        task & metric & scratch & all & charges & nmr & fukui\_n & fukui\_e & masking & homo-lumo \\
        \midrule
        caco2\_wang & MAE$\downarrow$ & 0.442 ± 0.041 & \textbf{0.354 ± 0.015} & \textbf{0.404 ± 0.069} & \textbf{0.364 ± 0.046} & \textbf{0.346 ± 0.034} & 0.483 ± 0.036 & 0.471 ± 0.080 & 0.381 ± 0.040 \\
        hia\_hou & ROC-AUC $\uparrow$ & \textbf{0.972 ± 0.015} & \textbf{0.982 ± 0.003} & \textbf{0.973 ± 0.027} & \textbf{0.977 ± 0.011} & 0.967 ± 0.011 & 0.908 ± 0.019 & \textbf{0.981 ± 0.013} & 0.869 ± 0.037 \\
        pgp\_broccatelli& ROC-AUC $\uparrow$  & 0.892 ± 0.011 & \textbf{0.913 ± 0.015} & 0.902 ± 0.019 & \textbf{0.917 ± 0.009} & 0.896 ± 0.020 & \textbf{0.911 ± 0.008} & \textbf{0.921 ± 0.003} & 0.870 ± 0.016 \\
        bioavailability\_ma& ROC-AUC $\uparrow$  & 0.606 ± 0.040 & \textbf{0.673 ± 0.028} & \textbf{0.662 ± 0.071} & 0.640 ± 0.040 & 0.663 ± 0.025 & 0.616 ± 0.082 & \textbf{0.698 ± 0.035} & \textbf{0.667 ± 0.031} \\
        lipophilicity\_astrazeneca& MAE $\downarrow$  & 0.539 ± 0.036 & \textbf{0.393 ± 0.005}$^*$ & \textbf{0.425 ± 0.023}$^*$ & 0.424 ± 0.007$^*$ & 0.457 ± 0.008$^*$ & 0.463 ± 0.011$^*$ & 0.462 ± 0.005$^*$ & 0.451 ± 0.011$^*$ \\
        solubility\_aqsoldb & MAE $\downarrow$  & 0.878 ± 0.031 & \textbf{0.720 ± 0.010}$^*$ & 0.726 ± 0.011$^*$ & 0.728 ± 0.014$^*$ & 0.756 ± 0.012 & 0.771 ± 0.015 & 0.769 ± 0.007 & 0.772 ± 0.019 \\
        bbb\_martins & ROC-AUC $\uparrow$  & 0.860 ± 0.016 & \textbf{0.872 ± 0.021} & \textbf{0.874 ± 0.011} & \textbf{0.869 ± 0.014} & 0.848 ± 0.018 & 0.845 ± 0.014 & \textbf{0.861 ± 0.025} & \textbf{0.883 ± 0.007} \\
        ppbr\_az & MAE $\downarrow$  & 8.477 ± 0.483 & \textbf{7.589 ± 0.203} & \textbf{7.668 ± 0.236} & \textbf{7.542 ± 0.215} & \textbf{7.530 ± 0.318} & 8.026 ± 0.222 & \textbf{8.056 ± 0.340} & \textbf{7.874 ± 0.287} \\
        vdss\_lombardo & Spearman $\uparrow$  & 0.554 ± 0.049 & 0.624 ± 0.020 & \textbf{0.637 ± 0.022} & 0.616 ± 0.034 & 0.616 ± 0.015 & \textbf{0.652 ± 0.012} & 0.620 ± 0.023 & 0.580 ± 0.029 \\
        cyp2d6\_veith & PR-AUC $\uparrow$  & 0.549 ± 0.043 & \textbf{0.621 ± 0.046} & \textbf{0.675 ± 0.014} & \textbf{0.643 ± 0.036} & 0.660 ± 0.009 & 0.638 ± 0.011 & 0.612 ± 0.021 & 0.612 ± 0.028 \\
        cyp3a4\_veith & PR-AUC $\uparrow$ & 0.799 ± 0.012 & 0.797 ± 0.029 & \textbf{0.847 ± 0.022} & \textbf{0.824 ± 0.021} & \textbf{0.838 ± 0.016} & \textbf{0.828 ± 0.018} & 0.817 ± 0.014 & 0.794 ± 0.018 \\
        cyp2c9\_veith & PR-AUC $\uparrow$ & 0.706 ± 0.014 & 0.703 ± 0.022 & \textbf{0.726 ± 0.024} & \textbf{0.739 ± 0.011} & \textbf{0.722 ± 0.021} & \textbf{0.734 ± 0.014} & \textbf{0.736 ± 0.014} & 0.708 ± 0.010 \\
        cyp2d6\_substrate\_carbonmangels & PR-AUC $\uparrow$ & 0.546 ± 0.042 & \textbf{0.648 ± 0.031} & \textbf{0.634 ± 0.050} & \textbf{0.653 ± 0.023} & \textbf{0.619 ± 0.057} & 0.578 ± 0.052 & \textbf{0.677 ± 0.022} & 0.582 ± 0.036 \\
        cyp3a4\_substrate\_carbonmangels & ROC-AUC $\uparrow$ & 0.637 ± 0.027 & 0.630 ± 0.015 & 0.646 ± 0.020 & 0.642 ± 0.009 & 0.645 ± 0.015 & 0.635 ± 0.031 & 0.641 ± 0.030 & \textbf{0.685 ± 0.015} \\
        cyp2c9\_substrate\_carbonmangels & PR-AUC $\uparrow$ & 0.360 ± 0.022 & 0.374 ± 0.028 & \textbf{0.404 ± 0.027} & 0.394 ± 0.024 & \textbf{0.405 ± 0.036} & 0.375 ± 0.030 & \textbf{0.396 ± 0.024} & \textbf{0.439 ± 0.043} \\
        half\_life\_obach & Spearman $\uparrow$ & 0.373 ± 0.076 & 0.462 ± 0.154 & \textbf{0.559 ± 0.034} & 0.487 ± 0.045 & 0.486 ± 0.030 & 0.476 ± 0.015 & 0.462 ± 0.052 & 0.426 ± 0.039 \\
        clearance\_microsome\_az & Spearman $\uparrow$ & 0.448 ± 0.038 & 0.548 ± 0.029 & \textbf{0.620 ± 0.007} & \textbf{0.613 ± 0.014} & 0.554 ± 0.019 & 0.513 ± 0.022 & 0.555 ± 0.022 & 0.565 ± 0.032 \\
        clearance\_hepatocyte\_az & Spearman $\uparrow$ & 0.336 ± 0.050 & 0.382 ± 0.032 & \textbf{0.456 ± 0.015} & 0.460 ± 0.019 & 0.374 ± 0.021 & 0.353 ± 0.028 & \textbf{0.478 ± 0.018} & 0.413 ± 0.030 \\
        herg & ROC-AUC $\uparrow$ & 0.709 ± 0.080 & 0.788 ± 0.029 & 0.824 ± 0.046 & 0.834 ± 0.030 & 0.752 ± 0.042 & 0.758 ± 0.053 & \textbf{0.880 ± 0.003} & 0.790 ± 0.031 \\
        ames & ROC-AUC $\uparrow$ & 0.772 ± 0.022 & \textbf{0.822 ± 0.005} & \textbf{0.821 ± 0.010} & \textbf{0.833 ± 0.014} & \textbf{0.820 ± 0.009} & \textbf{0.823 ± 0.012} & 0.801 ± 0.008 & 0.808 ± 0.008 \\
        dili & ROC-AUC $\uparrow$ & 0.856 ± 0.037 & \textbf{0.892 ± 0.033} & \textbf{0.859 ± 0.055} & \textbf{0.898 ± 0.022} & 0.847 ± 0.016 & 0.812 ± 0.122 & \textbf{0.906 ± 0.021} & 0.854 ± 0.017 \\
        ld50\_zhu & MAE $\downarrow$ & 0.593 ± 0.038 & 0.559 ± 0.016 & 0.571 ± 0.012 & \textbf{0.538 ± 0.014}$^*$ & 0.592 ± 0.029 & 0.618 ± 0.014 & 0.577 ± 0.010 & 0.582 ± 0.031 \\
        \hline
        Number of best models &\rule{0pt}{2.5ex} & 1 & 12 & 17 & 13 & 7 & 5 & 11 & 5 \\
        \bottomrule
    \end{tabular}%
}
\end{table}

\paragraph{\textbf{Performance on internal microsomal clearance data}}
Although the TDC dataset provides a well established benchmark in modeling ADMET properties, the different models reported here demonstrated close performance on multiple taks. Expecting divergence of model metrics, we tested our methodology on a much larger dataset of proprietary JNJ HLM clearance data and summarized the results in table \ref{clearance_micro}. The models pretrained on all atomic QM properties obtain the best results in both metrics ($R^2$ and Spearman's coefficient), followed closely by models pretrained on NMR shifts and atomic charges. Models pretrained on Fukui indices give the lowest results among models pretrained on atomic QM properties, obtaining similar performances to models pretrained on HLG. Notably, and contrary to what seen in the benchmark results, models pretrained using masking obtain the worst results over all pretrained models, albeit still giving improvements over models trained from scratch.

\begin{table} 
\caption{Results of the fine-tuning on internal microsomal clearance dataset. Results are reported for both values of clearance in the dataset and for all pretraining strategies both in terms of $R^2$ coefficient and in terms of Spearman's rank coefficient.}
\label{clearance_micro}
\begin{center}
\begin{small}
\begin{sc}
\resizebox{\textwidth}{!}{
    \begin{tabular}{llllllllll}
\toprule
& metric & scratch & all & charges & nmr & fukui\_n & fukui\_e & masking & homo-lumo \\
\midrule
clearance\_1 & $R^2$ $\uparrow$ & 0.505 ± 0.010 & 0.640 ± 0.004 & 0.629 ± 0.006 & 0.635 ± 0.006 & 0.599 ± 0.004 & 0.593 ± 0.004 & 0.580 ± 0.012 & 0.602 ± 0.006 \\
 & Spearman $\uparrow$ & 0.728 ± 0.008 & 0.807 ± 0.003 & 0.799 ± 0.004 & 0.801 ± 0.003 & 0.785 ± 0.003 & 0.785 ± 0.001 & 0.774 ± 0.007 & 0.786 ± 0.004 \\
clearance\_2 & $R^2$ $\uparrow$ & 0.534 ± 0.006 & 0.653 ± 0.004 & 0.633 ± 0.003 & 0.643 ± 0.005 & 0.598 ± 0.007 & 0.610 ± 0.008 & 0.597 ± 0.002 & 0.607 ± 0.005 \\
 & Spearman $\uparrow$ & 0.750 ± 0.005 & 0.818 ± 0.003 & 0.807 ± 0.004 & 0.811 ± 0.002 & 0.789 ± 0.002 & 0.795 ± 0.006 & 0.786 ± 0.002 & 0.794 ± 0.002 \\
\bottomrule
\end{tabular}%
    }
    \end{sc}
    \end{small}
    \end{center}
\end{table}

\begin{figure}
    \centering
    \includegraphics[width = \textwidth]{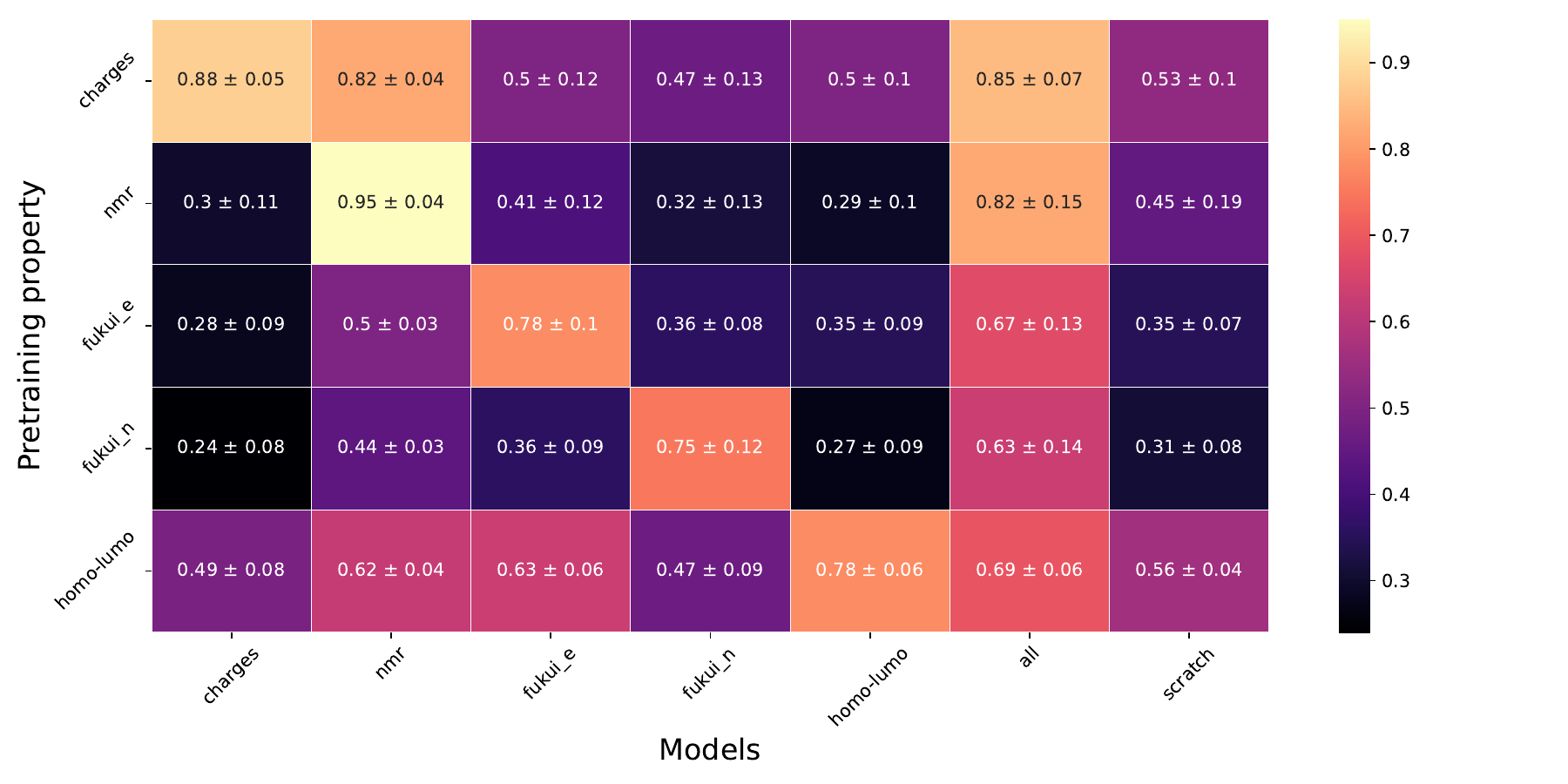}
    \caption{$R^2$ for the regression tasks using the representations of a sample of the pretraining data obtained with fine-tuned models. We report the mean and standard deviation over all fine-tuning cases (mean and standard deviation over twenty-two cases).}
    \label{fig:crossr2}
\end{figure}

\paragraph{\textbf{Conservation of pretraining information after fine-tuning}}
The results obtained on the regularized linear regression of pretraining labels from the representations of the pretraining structures obtained with fine-tuned models are reported in Fig. \ref{fig:crossr2}. We report each value of $R^2$ coefficient with mean and standard deviation over the results obtained from the twenty-two fine-tuned models obtained from each pretraining (masking is excluded from this analysis as it is self-supervised). The obtained representations are found to maintain a high degree of linear correlation with their correspondent pretraining property even after being fine-tuned on downstream tasks both in absolute terms and with respect to the model trained from scratch. In particular, models pretrained on NMR shifts seems to preserve its pretraining information the most, providing good representations also for the regression of atomic charges. On the other hand, models pretrained on Fukui function values and HLG have slightly less linear correlation with the pretraining task, although still being in a quite high range when considering standard deviation and the comparison with the models trained from scratch. Models pretrained on all four atomic properties maintain a high degree of linear correlation with individual properties, but especially with charges and NMR shifts. Finally, we notice the models pretrained on NMR shifts, Fukui electrophilic indices and all atomic properties, exhibit some degree of correlation with HLGs, but not the other way around: models pretrained on HLG do not provide strong correlation with any atom-level property. We hypothesize that the way the network is trained might be responsible for this asymmetric behaviour. Training on atomic properties followed by fine-tuning on downstream molecular properties impacts in the final layer both the lantent representations relative to the atoms and the one relative to the CLS token used to model molecular properties. This is not the case for HLGs pretrained models, as in the final layer only the representation associated with CLS tokens are directly used to compute the output value.

\begin{figure}
    \centering
    \includegraphics[width = \textwidth]{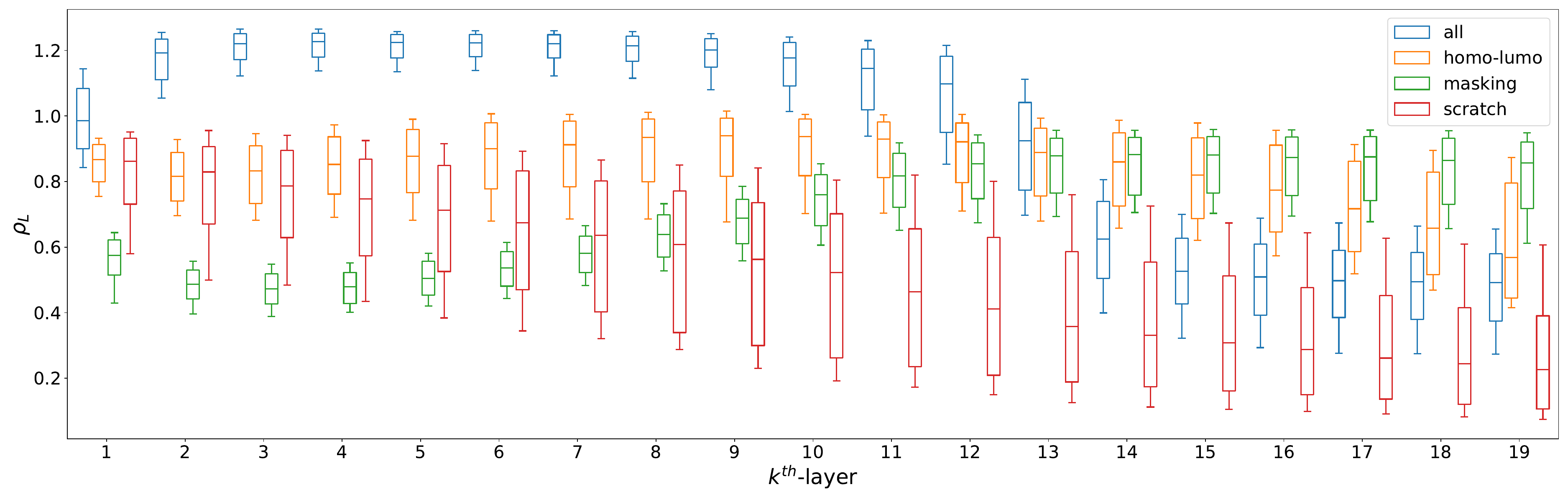}
    \caption{Expressivity of the latent representation measured with the quantity $\rho_L$ as a function of layer number. This quantity is computed for a sample of 2200 structures extracted uniformly from all the fine-tuning test sets (100 structures for each of the 22 tasks) and results are reported as boxplots at each layer. This is done for models pretrained on HLG, models pretrained on all atom-level QM properties, models pretrained with masking and models trained from scratch. The whiskers go from the 15th percentile to the 85th for better visualization of trends and outliers are excluded for the same reason.}
    \label{fig:rank_collapse}
\end{figure}

\paragraph{\textbf{Latent expressivity as across layers}} 
The results of the analysis of $\rho_L$ across layers are summarized for the models using the three main pretraining strategies (all atom-level quantum properties, HLG and masking) and for the models trained from scratch in Fig. \ref{fig:rank_collapse}, while a similar plot comparing the models pretrained on each atom-level QM property separately is reported in the SI. It is evident from the plot that the trend of $\rho_L$ is very different across pretraining methods.
Overall, when comparing to the models trained from scratch, all pretraining strategies mitigate the collapse in latent expressivity. In particular, while models pretrained on HLG are characterized by a constant level of expressivity across layers, models pretrained using masking have more similar atomic latent representations in the first few layers and more dissimilar in the last ones. For models pretrained on all atomic quantum properties, a strong increase in expressivity is observed in the first part of the network, reaching a higher value of $\rho_L$ than the other cases, followed by a decrease in the last part, closer to the regression head. Regarding pretraining on individual atomic properties, we find that the NMR shifts and charges models have similar behaviour. The ones pretrained on Fukui indices, while presenting a similar trend, achieve a lower maximum expressivity more similar to the models pretrained on HLG. The absence of complete expressivity collapse for pretrained models is likely comes from the much higher number of examples that the models were trained on comparably to the models trained from scratch. However, dissimilar behaviours of expressivity indicate that the pretraining strategies explored here produce very different models even when the overall performance improvements on the benchmark are comparable. It is notable that if we consider the highest value across layers for each model, models pretrained on atomic quantum properties achieve the highest maximum latent expressivity. While we do not have a definitive explaination for the final sharp collapse in the last layers, we hypothesise that quantum atomic property regression requires the model to capture correlations between atoms within the same molecular structure when close to the last layer, as these properties depend on the surrounding atoms as do their respective properties. This would be sustained from the seemingly opposite trend found in the models pretrained with masking, which is a classification task that requires to maximally distinguish atomic latent representations close to the last network layer using a cross entropy loss function that rewards higher certainties.

\begin{figure}
    \centering
    \includegraphics[width=\textwidth]{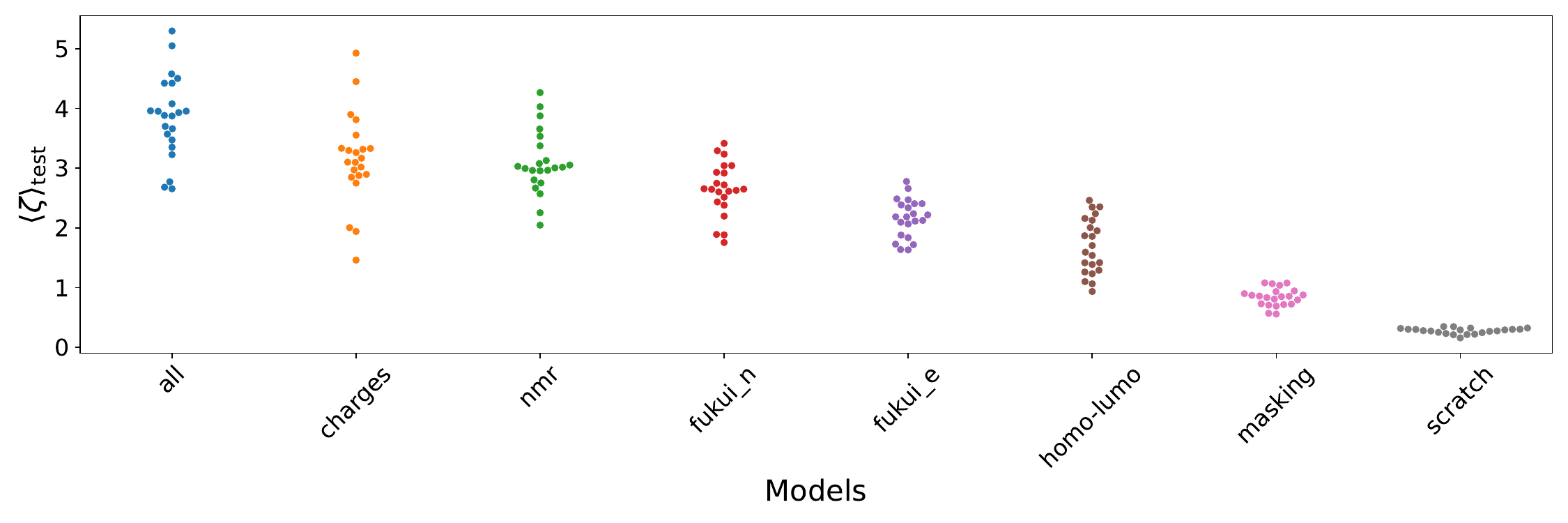}
    \caption{Spectral perception of the input graphs for the models fine-tuned on the TDC datasets grouped by pretraining strategy. This is reported in the form of swarm plots of the values of $\zeta$ averaged across each of the 22 fine-tuning test sets for fixed pretraining strategy.}
    \label{fig: zetaviolin}
\end{figure}

\paragraph{\textbf{Spectral Analysis of Attention Rollout}}
We evaluate the metric $\zeta$ defined in Eq. \ref{zeta_metric} as described in the Methods section obtaining a distribution of 22 values over the downstream tasks per each group of studied models. The results are reported in Fig. \ref{fig: zetaviolin} as a set of swarm plots. Firstly, it is evident that the models trained from scratch present values of $\zeta$ that are close to $0$ indicating little to no presence of non-trivial Laplacian eigenmodes in the spectrum of their $\Tilde{A}$ matrix. On the contrary, every pretrained model (including masking) presents nonzero values of $\zeta$ across the downstream tasks raging from $\sim 1$ to $\sim 6$. Within this last group of models one can clearly notice how pretraining on the atom-level QM properties provides the strongest increase of perception of graph Laplacian eigenmodes. In particular, the model pretrained using all properties in a multi-task fashion presents the highest values of $\zeta$, followed by the models pretrained on charges, NMR shifts, nucleophilic and electrophilic Fukui function indices. The models pretrained on HLGs also present some degree of spectral perception, albeit in a lower range than the previously mentioned models, followed by models pretrained using masking which present the lowest graph spectral perception among the set of pretrained models. 

\begin{figure}
    \centering
    \includegraphics[width=\textwidth]{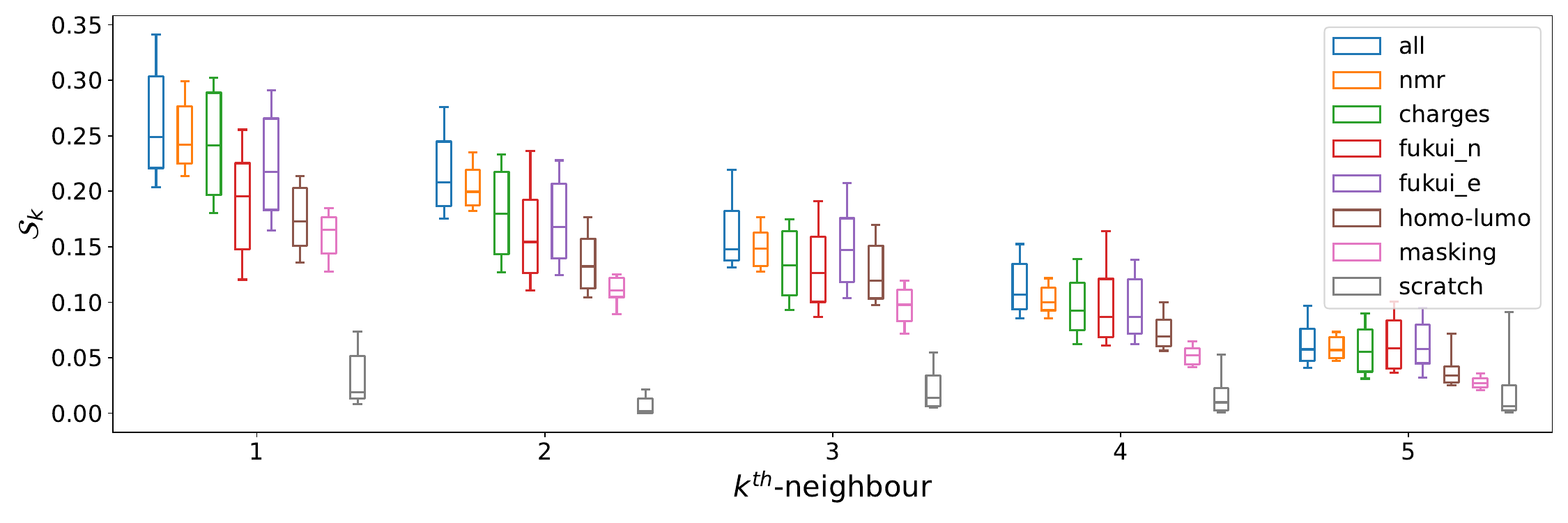}
    \caption{Boxplots of the $k^{th}$ neighbour normalized sensitivities $\mathcal{S}_k$ for $k\in [1, \dots,5]$. Each boxplot summarizes a sample of 1100 structures extracted uniformly from all the fine-tuning test sets (50 structures for each of the 22 tasks). We report this quantity for all studied pretraining strategies, and also for the models trained from scratch. The whiskers cover the values from the 15th percentile to the 85th for better visualization of trends. Outliers are excluded for the same reason.}
    \label{fig:oversquashing}
\end{figure}

\paragraph{\textbf{Neighbour sensitivity analysis}}
The results of the neighbour sensitivity analysis are reported in Fig. \ref{fig:oversquashing}. For each considered group of models we report the value of $\mathcal{S}_k$ for $k\in[1, \dots , 5]$ in boxplots over 1100 structures sampled uniformly from the test sets of the fine-tuning tasks. It is found that the models trained from scratch exhibits a constant and low sensitivity of representation with respect to neighbouring atoms, whereas pretrained models present a reasonable descending trend with topological distance. In particular, the models pretrained on all the atomic QM properties have a stronger sensitivity than all other models for all the considered topological distances, especially, for first and second neighbours. The models pretrained on HLG present slightly higher sensitivities than the ones pretrained using masking which presents the lowest set of sensitivities among all pretraining strategies. Among the models trained on individual atomic QM properties, the sensitivity ranges tend to overlap, but they are positioned in between the models pretrained on HLGs and models pretrained on all atomic QM properties for all considered topological distances.

\section{Discussion}

Our in-depth analysis demonstrates that, among the tested strategies, pretraining the Graphormer on four atomic QM properties in the multitask fashion provides the best model for subsequent fine-tuning on ADMET properties. The final models exhibit high performance results in the TDC benchmark and, more importantly, outperform other models on a much larger dataset of JNJ HLM clearance data. The latent space analysis also positions the respective models at the top with highest values of latent expressivity, neighbour sensitivities, and graph-spectral perception. Besides, last layer representations of the respective models pretrained on four atomic properties retain high degree of correlation with all types of pretraining atomic data after fine-tuning. 

Pretraining on NMR shielding constants and atomic charges yields the models that are sharing the second overall rank in studied metrics. Moreover, these pretrainings also provide the highest number of top-performant models in the TDC dataset. Interestingly, while pretraining charge labels can be modeled with good results using NMR-pretrained models after dowsntream tasks fine-tuning, the opposite doesn't hold, which indicates that NMR shifts may contain richer information than atomic charges. NMR chemical shifts are indeed known for their extreme sensitivity to the atomic environment of the respective nuclei, covering both electronic and to some extent steric effects.

The Fukui indices pretraining posses the third cumulative rank among the studied approaches. Unexpectedly, the respective models were not among the top performing models for clearance modelling, despite stong relevance of electrophilic Fukui indices to site of metabolism predictions\cite{Beck2005} which in turn drives the hepatic clearance of drugs. Fukui indices are often calculated with the aim to build affinity QSAR models, particularly to CYP enzymes \cite{VanDamme2009}. This observation suggests that there is more to pretraining such complex models than only transfer learning between tasks.

Models pretrained on HLG resulted being close to models pretrained on Fukui indices under the performance indicators on both TDC and JNJ HLM data, and just slightly worse under representation indicators. While the relevance of HLG to ADMET properties is arguable, it was used as a feature for QSAR modelling of CYP enzyme inhibition\cite{Lewis1997, Ai2010} and generally characterizes the propensity of a molecule to donate or accept electrons, sometimes referred as a global hardness \cite{Kaya2015}. In particular, Fukui indices and frontier orbital levels are tightly related properties and often calculated to characterize biologically active molecules. 
When comparing HLG pretraining to Fukui indices pretraining, though, it is important to notice that the atomic QM properties dataset contains $\sim$ 20 times less molecular structures than the HLG pretraining dataset, making atomic properties much more efficient in terms of training time and resources. At the same time, because each molecule in the atomic properties pretraining dataset contains on average 17 non-hydrogen atoms per molecule, the overall number of atomic numerical properties is on the same scale as the HLG dataset. Furthermore, because calculation of atomic properties typically requires only a fraction of overall computational resources spent on geometry optimization and electronic structure refinement during QM modeling, such properties provide a finer grade physical description of molecular structures with a non-dramatic overhead in the data generation phase. 

Masking pretraining, considered as example of label-free atom-level pretraining, provided inconsistent results. If the performance on the TDC benchmark is comparable with other models pretrained on atomic QM properties, the analysis of the latent representations together with the results on the much wider JNJ HLM dataset place this pretraining strategy at the bottom among the tested ones, confirming that the improvement seen with atom-level QM pretraining does not come solely from being atom-resolved. It is worth noting that for consistency we utilized the same set of molecules for the masking pretraining as for the pretraining on atomic QM properties. Considering simplicity of data preparation, masking pretraining can be used with much larger datasets containing tens of millions of molecules and potentially improve the performance of the respective models, however such experiments were beyond the scope of the present study.

These last findings furthermore highlight the limitation of picking the best pretraining method solely using the results obtained on public benchmark datasets and the importance of utilizing other metrics. In this regard, we would like to emphasize that novel analysis of the spectrum of the attention rollout matrix documented a non-trivial effect arising in pretrained GTs where the model, albeit under the strong approximation of the explainability method, shows hints of filtered spectral graph convolution. Such findings connect the GT architecture to the family of Spectral Graph Neural Networks (SGNNs) \cite{lecunnspectral,wellingspectral, waveletspectral, fastspectral}, and should stimulate further research potentially leading to the development of more robust models that can better leverage both on the graph-spectral features typical of SGNNs and the flexibility of transformer-based architectures for graph-based applications.

We hope that the present work provides a different perspective on model evaluation for ADMET modeling tasks, as well as valuable insights for future research in molecular representation learning and for the development of useful in-silico datasets.

\section{Declarations}

\subsection{Funding}
This research was financially supported by the European Union’s Horizon 2020 research and innovation program under the Marie Skłodowska-Curie grant agreement No 956832, “Advanced Machine learning for Innovative Drug Discovery” (AIDD).

\subsection{Acknowledgements}
We thank Dr. Leonardo Medrano Sandonas for the helpful comments. A.F. thanks Dr. Matthieu Sarkis for the discussions around mathematical properties of graph Laplacians.

\subsection{Availability of data and materials}
The code pertaining the results on public data will be made available on \href{https://github.com/aidd-msca/GraphQPT}{https://github.com/aidd-msca/GraphQPT}.

\subsection{Preprint statement}
This article is a preprint reporting new medical research and has not been peer-reviewed; it should not be used to guide clinical practice or be reported in the press as conclusive.

\subsection{Authors' contributions}
A.F. wrote the code for the paper results, conducted model training and evaluation, devised and performed all representation analyses. R.N. conceived the idea of atom-level quantum mechanical properties pretraining, developed the chytorch version of Graphormer, curated datasets, and provided code support and deep learning advice. J.A.-M. proposed the use of the TDC benchmark, contributed to preliminary pretraining and fine-tuning results, offered deep learning and code guidance. K.C. contributed to the attention rollout spectral analysis and supervised all stages of the project. All authors actively discussed the results and contributed to the final manuscript.

\bibliography{bibliografia}

\providecommand{\latin}[1]{#1}
\makeatletter
\providecommand{\doi}
  {\begingroup\let\do\@makeother\dospecials
  \catcode`\{=1 \catcode`\}=2 \doi@aux}
\providecommand{\doi@aux}[1]{\endgroup\texttt{#1}}
\makeatother
\providecommand*\mcitethebibliography{\thebibliography}
\csname @ifundefined\endcsname{endmcitethebibliography}  {\let\endmcitethebibliography\endthebibliography}{}
\begin{mcitethebibliography}{76}
\providecommand*\natexlab[1]{#1}
\providecommand*\mciteSetBstSublistMode[1]{}
\providecommand*\mciteSetBstMaxWidthForm[2]{}
\providecommand*\mciteBstWouldAddEndPuncttrue
  {\def\EndOfBibitem{\unskip.}}
\providecommand*\mciteBstWouldAddEndPunctfalse
  {\let\EndOfBibitem\relax}
\providecommand*\mciteSetBstMidEndSepPunct[3]{}
\providecommand*\mciteSetBstSublistLabelBeginEnd[3]{}
\providecommand*\EndOfBibitem{}
\mciteSetBstSublistMode{f}
\mciteSetBstMaxWidthForm{subitem}{(\alph{mcitesubitemcount})}
\mciteSetBstSublistLabelBeginEnd
  {\mcitemaxwidthsubitemform\space}
  {\relax}
  {\relax}

\bibitem[Rupp \latin{et~al.}(2012)Rupp, Tkatchenko, M\"uller, and von Lilienfeld]{coulombmatrix}
Rupp,~M.; Tkatchenko,~A.; M\"uller,~K.-R.; von Lilienfeld,~O.~A. Fast and Accurate Modeling of Molecular Atomization Energies with Machine Learning. \emph{Phys. Rev. Lett.} \textbf{2012}, \emph{108}, 058301\relax
\mciteBstWouldAddEndPuncttrue
\mciteSetBstMidEndSepPunct{\mcitedefaultmidpunct}
{\mcitedefaultendpunct}{\mcitedefaultseppunct}\relax
\EndOfBibitem
\bibitem[Hansen \latin{et~al.}(2015)Hansen, Biegler, Ramakrishnan, Pronobis, von Lilienfeld, Müller, and Tkatchenko]{BoB}
Hansen,~K.; Biegler,~F.; Ramakrishnan,~R.; Pronobis,~W.; von Lilienfeld,~O.~A.; Müller,~K.-R.; Tkatchenko,~A. Machine Learning Predictions of Molecular Properties: Accurate Many-Body Potentials and Nonlocality in Chemical Space. \emph{The Journal of Physical Chemistry Letters} \textbf{2015}, \emph{6}, 2326--2331, PMID: 26113956\relax
\mciteBstWouldAddEndPuncttrue
\mciteSetBstMidEndSepPunct{\mcitedefaultmidpunct}
{\mcitedefaultendpunct}{\mcitedefaultseppunct}\relax
\EndOfBibitem
\bibitem[Huang \latin{et~al.}(2020)Huang, Symonds, and von Lilienfeld]{SLATM}
Huang,~B.; Symonds,~N.~O.; von Lilienfeld,~O.~A. In \emph{Handbook of Materials Modeling: Methods: Theory and Modeling}; Andreoni,~W., Yip,~S., Eds.; Springer International Publishing: Cham, 2020; pp 1883--1909\relax
\mciteBstWouldAddEndPuncttrue
\mciteSetBstMidEndSepPunct{\mcitedefaultmidpunct}
{\mcitedefaultendpunct}{\mcitedefaultseppunct}\relax
\EndOfBibitem
\bibitem[Chen \latin{et~al.}(2018)Chen, Engkvist, Wang, Olivecrona, and Blaschke]{Chen2018}
Chen,~H.; Engkvist,~O.; Wang,~Y.; Olivecrona,~M.; Blaschke,~T. The rise of deep learning in drug discovery. \emph{Drug Discovery Today} \textbf{2018}, \emph{23}, 1241--1250\relax
\mciteBstWouldAddEndPuncttrue
\mciteSetBstMidEndSepPunct{\mcitedefaultmidpunct}
{\mcitedefaultendpunct}{\mcitedefaultseppunct}\relax
\EndOfBibitem
\bibitem[Jayatunga \latin{et~al.}(2022)Jayatunga, Xie, Ruder, Schulze, and Meier]{Jayatunga2022}
Jayatunga,~M.~K.; Xie,~W.; Ruder,~L.; Schulze,~U.; Meier,~C. AI in Small-Molecule Drug Discovery: A Coming Wave. \emph{Nat. Rev. Drug Discov.} \textbf{2022}, \emph{21}, 175--176\relax
\mciteBstWouldAddEndPuncttrue
\mciteSetBstMidEndSepPunct{\mcitedefaultmidpunct}
{\mcitedefaultendpunct}{\mcitedefaultseppunct}\relax
\EndOfBibitem
\bibitem[Bule \latin{et~al.}(2021)Bule, Jalalimanesh, Bayrami, Baeeri, and Abdollahi]{Bule2021}
Bule,~M.; Jalalimanesh,~N.; Bayrami,~Z.; Baeeri,~M.; Abdollahi,~M. The rise of deep learning and transformations in bioactivity prediction power of molecular modeling tools. \emph{Chemical Biology \& Drug Design} \textbf{2021}, \emph{98}, 954--967\relax
\mciteBstWouldAddEndPuncttrue
\mciteSetBstMidEndSepPunct{\mcitedefaultmidpunct}
{\mcitedefaultendpunct}{\mcitedefaultseppunct}\relax
\EndOfBibitem
\bibitem[Li \latin{et~al.}(2022)Li, Huang, and Zitnik]{Li2022}
Li,~M.~M.; Huang,~K.; Zitnik,~M. Graph representation learning in biomedicine and healthcare. \emph{Nature Biomedical Engineering} \textbf{2022}, \emph{6}, 1353--1369\relax
\mciteBstWouldAddEndPuncttrue
\mciteSetBstMidEndSepPunct{\mcitedefaultmidpunct}
{\mcitedefaultendpunct}{\mcitedefaultseppunct}\relax
\EndOfBibitem
\bibitem[Chuang \latin{et~al.}(2020)Chuang, Gunsalus, and Keiser]{Chuang2020}
Chuang,~K.~V.; Gunsalus,~L.~M.; Keiser,~M.~J. Learning Molecular Representations for Medicinal Chemistry. \emph{Journal of Medicinal Chemistry} \textbf{2020}, \emph{63}, 8705--8722, PMID: 32366098\relax
\mciteBstWouldAddEndPuncttrue
\mciteSetBstMidEndSepPunct{\mcitedefaultmidpunct}
{\mcitedefaultendpunct}{\mcitedefaultseppunct}\relax
\EndOfBibitem
\bibitem[Born \latin{et~al.}(2023)Born, Markert, Janakarajan, Kimber, Volkamer, Martínez, and Manica]{Born2023}
Born,~J.; Markert,~G.; Janakarajan,~N.; Kimber,~T.~B.; Volkamer,~A.; Martínez,~M.~R.; Manica,~M. Chemical representation learning for toxicity prediction. \emph{Digital Discovery} \textbf{2023}, \emph{2}, 674--691\relax
\mciteBstWouldAddEndPuncttrue
\mciteSetBstMidEndSepPunct{\mcitedefaultmidpunct}
{\mcitedefaultendpunct}{\mcitedefaultseppunct}\relax
\EndOfBibitem
\bibitem[Wang \latin{et~al.}(2022)Wang, Wang, Cao, and Barati~Farimani]{Wang2022}
Wang,~Y.; Wang,~J.; Cao,~Z.; Barati~Farimani,~A. Molecular contrastive learning of representations via graph neural networks. \emph{Nature Machine Intelligence} \textbf{2022}, \emph{4}, 279--287\relax
\mciteBstWouldAddEndPuncttrue
\mciteSetBstMidEndSepPunct{\mcitedefaultmidpunct}
{\mcitedefaultendpunct}{\mcitedefaultseppunct}\relax
\EndOfBibitem
\bibitem[Kaufman \latin{et~al.}(2024)Kaufman, Williams, Underkoffler, Pederson, Mardirossian, Watson, and Parkhill]{Kaufman2024}
Kaufman,~B.; Williams,~E.~C.; Underkoffler,~C.; Pederson,~R.; Mardirossian,~N.; Watson,~I.; Parkhill,~J. COATI: Multimodal Contrastive Pretraining for Representing and Traversing Chemical Space. \emph{Journal of Chemical Information and Modeling} \textbf{2024}, \emph{64}, 1145--1157, PMID: 38316665\relax
\mciteBstWouldAddEndPuncttrue
\mciteSetBstMidEndSepPunct{\mcitedefaultmidpunct}
{\mcitedefaultendpunct}{\mcitedefaultseppunct}\relax
\EndOfBibitem
\bibitem[Ilnicka and Schneider(2023)Ilnicka, and Schneider]{encoderfingerprints}
Ilnicka,~A.; Schneider,~G. Compression of molecular fingerprints with autoencoder networks. \emph{Molecular Informatics} \textbf{2023}, \emph{42}, 2300059\relax
\mciteBstWouldAddEndPuncttrue
\mciteSetBstMidEndSepPunct{\mcitedefaultmidpunct}
{\mcitedefaultendpunct}{\mcitedefaultseppunct}\relax
\EndOfBibitem
\bibitem[Sanchez-Fernandez \latin{et~al.}(2022)Sanchez-Fernandez, Rumetshofer, Hochreiter, and Klambauer]{contrastiveimage}
Sanchez-Fernandez,~A.; Rumetshofer,~E.; Hochreiter,~S.; Klambauer,~G. Contrastive learning of image- and structure-based representations in drug discovery. ICLR2022 Machine Learning for Drug Discovery. 2022\relax
\mciteBstWouldAddEndPuncttrue
\mciteSetBstMidEndSepPunct{\mcitedefaultmidpunct}
{\mcitedefaultendpunct}{\mcitedefaultseppunct}\relax
\EndOfBibitem
\bibitem[Fang \latin{et~al.}(2023)Fang, Zhang, Zhang, Chen, Zhuang, Shao, Fan, and Chen]{KGcontrastive}
Fang,~Y.; Zhang,~Q.; Zhang,~N.; Chen,~Z.; Zhuang,~X.; Shao,~X.; Fan,~X.; Chen,~H. Knowledge graph-enhanced molecular contrastive learning with functional prompt. \emph{Nature Machine Intelligence} \textbf{2023}, \emph{5}, 542--553\relax
\mciteBstWouldAddEndPuncttrue
\mciteSetBstMidEndSepPunct{\mcitedefaultmidpunct}
{\mcitedefaultendpunct}{\mcitedefaultseppunct}\relax
\EndOfBibitem
\bibitem[Wen \latin{et~al.}(2023)Wen, Zhang, Rush, Panickan, Li, Cai, Zhou, Ho, Costa, Begoli, Hong, Gaziano, Cho, Lu, Liao, Zitnik, and Cai]{medicalrecords}
Wen,~J. \latin{et~al.}  {Multimodal representation learning for predicting molecule–disease relations}. \emph{Bioinformatics} \textbf{2023}, \emph{39}, btad085\relax
\mciteBstWouldAddEndPuncttrue
\mciteSetBstMidEndSepPunct{\mcitedefaultmidpunct}
{\mcitedefaultendpunct}{\mcitedefaultseppunct}\relax
\EndOfBibitem
\bibitem[Su \latin{et~al.}(2022)Su, Du, Yang, Zhou, Li, Rao, Sun, Lu, and Wen]{nlpmultimodal}
Su,~B.; Du,~D.; Yang,~Z.; Zhou,~Y.; Li,~J.; Rao,~A.; Sun,~H.; Lu,~Z.; Wen,~J.-R. A Molecular Multimodal Foundation Model Associating Molecule Graphs with Natural Language. 2022; \url{https://arxiv.org/abs/2209.05481}\relax
\mciteBstWouldAddEndPuncttrue
\mciteSetBstMidEndSepPunct{\mcitedefaultmidpunct}
{\mcitedefaultendpunct}{\mcitedefaultseppunct}\relax
\EndOfBibitem
\bibitem[Wang \latin{et~al.}(2024)Wang, Jiang, Wang, and Xuan]{seqgraphgeom}
Wang,~Z.; Jiang,~T.; Wang,~J.; Xuan,~Q. Multi-Modal Representation Learning for Molecular Property Prediction: Sequence, Graph, Geometry. 2024; \url{https://arxiv.org/abs/2401.03369}\relax
\mciteBstWouldAddEndPuncttrue
\mciteSetBstMidEndSepPunct{\mcitedefaultmidpunct}
{\mcitedefaultendpunct}{\mcitedefaultseppunct}\relax
\EndOfBibitem
\bibitem[Gao \latin{et~al.}(2020)Gao, Ramezanghorbani, Isayev, Smith, and Roitberg]{torchani}
Gao,~X.; Ramezanghorbani,~F.; Isayev,~O.; Smith,~J.~S.; Roitberg,~A.~E. TorchANI: A Free and Open Source PyTorch-Based Deep Learning Implementation of the ANI Neural Network Potentials. \emph{Journal of Chemical Information and Modeling} \textbf{2020}, \emph{60}, 3408--3415, PMID: 32568524\relax
\mciteBstWouldAddEndPuncttrue
\mciteSetBstMidEndSepPunct{\mcitedefaultmidpunct}
{\mcitedefaultendpunct}{\mcitedefaultseppunct}\relax
\EndOfBibitem
\bibitem[Sch\"{u}tt \latin{et~al.}(2017)Sch\"{u}tt, Kindermans, Sauceda~Felix, Chmiela, Tkatchenko, and M\"{u}ller]{schnet}
Sch\"{u}tt,~K.; Kindermans,~P.-J.; Sauceda~Felix,~H.~E.; Chmiela,~S.; Tkatchenko,~A.; M\"{u}ller,~K.-R. SchNet: A continuous-filter convolutional neural network for modeling quantum interactions. Advances in Neural Information Processing Systems. 2017\relax
\mciteBstWouldAddEndPuncttrue
\mciteSetBstMidEndSepPunct{\mcitedefaultmidpunct}
{\mcitedefaultendpunct}{\mcitedefaultseppunct}\relax
\EndOfBibitem
\bibitem[Batzner \latin{et~al.}(2022)Batzner, Musaelian, Sun, Geiger, Mailoa, Kornbluth, Molinari, Smidt, and Kozinsky]{nequip}
Batzner,~S.; Musaelian,~A.; Sun,~L.; Geiger,~M.; Mailoa,~J.~P.; Kornbluth,~M.; Molinari,~N.; Smidt,~T.~E.; Kozinsky,~B. E(3)-equivariant graph neural networks for data-efficient and accurate interatomic potentials. \emph{Nature Communications} \textbf{2022}, \emph{13}, 2453\relax
\mciteBstWouldAddEndPuncttrue
\mciteSetBstMidEndSepPunct{\mcitedefaultmidpunct}
{\mcitedefaultendpunct}{\mcitedefaultseppunct}\relax
\EndOfBibitem
\bibitem[Deng \latin{et~al.}(2023)Deng, Yang, and et~al.]{Deng:23}
Deng,~J.; Yang,~Z.; et~al.,~H.~W. A systematic study of key elements underlying molecular property prediction. \emph{Nat Commun} \textbf{2023}, \emph{14}, 6395\relax
\mciteBstWouldAddEndPuncttrue
\mciteSetBstMidEndSepPunct{\mcitedefaultmidpunct}
{\mcitedefaultendpunct}{\mcitedefaultseppunct}\relax
\EndOfBibitem
\bibitem[Dou \latin{et~al.}(2023)Dou, Zhu, Merkurjev, Ke, Chen, Jiang, Zhu, Liu, Zhang, and Wei]{Dourew2023}
Dou,~B.; Zhu,~Z.; Merkurjev,~E.; Ke,~L.; Chen,~L.; Jiang,~J.; Zhu,~Y.; Liu,~J.; Zhang,~B.; Wei,~G.-W. Machine Learning Methods for Small Data Challenges in Molecular Science. \emph{Chemical Reviews} \textbf{2023}, \emph{123}, 8736--8780, PMID: 37384816\relax
\mciteBstWouldAddEndPuncttrue
\mciteSetBstMidEndSepPunct{\mcitedefaultmidpunct}
{\mcitedefaultendpunct}{\mcitedefaultseppunct}\relax
\EndOfBibitem
\bibitem[Glavatskikh \latin{et~al.}(2019)Glavatskikh, Leguy, Hunault, Cauchy, and Da~Mota]{Glavatskikh2019}
Glavatskikh,~M.; Leguy,~J.; Hunault,~G.; Cauchy,~T.; Da~Mota,~B. Dataset’s chemical diversity limits the generalizability of machine learning predictions. \emph{Journal of Cheminformatics} \textbf{2019}, \emph{11}, 69\relax
\mciteBstWouldAddEndPuncttrue
\mciteSetBstMidEndSepPunct{\mcitedefaultmidpunct}
{\mcitedefaultendpunct}{\mcitedefaultseppunct}\relax
\EndOfBibitem
\bibitem[Ektefaie \latin{et~al.}(2024)Ektefaie, Shen, Bykova, Marin, Zitnik, and Farhat]{Ektefaie2024}
Ektefaie,~Y.; Shen,~A.; Bykova,~D.; Marin,~M.; Zitnik,~M.; Farhat,~M. Evaluating generalizability of artificial intelligence models for molecular datasets. \emph{bioRxiv} \textbf{2024}, \relax
\mciteBstWouldAddEndPunctfalse
\mciteSetBstMidEndSepPunct{\mcitedefaultmidpunct}
{}{\mcitedefaultseppunct}\relax
\EndOfBibitem
\bibitem[Broccatelli \latin{et~al.}(2022)Broccatelli, Trager, Reutlinger, Karypis, and Li]{Broccatelli2022}
Broccatelli,~F.; Trager,~R.; Reutlinger,~M.; Karypis,~G.; Li,~M. Benchmarking Accuracy and Generalizability of Four Graph Neural Networks Using Large In Vitro ADME Datasets from Different Chemical Spaces. \emph{Molecular Informatics} \textbf{2022}, \emph{41}, 2100321\relax
\mciteBstWouldAddEndPuncttrue
\mciteSetBstMidEndSepPunct{\mcitedefaultmidpunct}
{\mcitedefaultendpunct}{\mcitedefaultseppunct}\relax
\EndOfBibitem
\bibitem[David Z~Huang and Bahmanyar(2021)David Z~Huang, and Bahmanyar]{Huang2021}
David Z~Huang,~J. C.~B.; Bahmanyar,~S.~S. The challenges of generalizability in artificial intelligence for ADME/Tox endpoint and activity prediction. \emph{Expert Opinion on Drug Discovery} \textbf{2021}, \emph{16}, 1045--1056, PMID: 33739897\relax
\mciteBstWouldAddEndPuncttrue
\mciteSetBstMidEndSepPunct{\mcitedefaultmidpunct}
{\mcitedefaultendpunct}{\mcitedefaultseppunct}\relax
\EndOfBibitem
\bibitem[Keith \latin{et~al.}(2021)Keith, Vassilev-Galindo, Cheng, Chmiela, Gastegger, Müller, and Tkatchenko]{Valentinrew}
Keith,~J.~A.; Vassilev-Galindo,~V.; Cheng,~B.; Chmiela,~S.; Gastegger,~M.; Müller,~K.-R.; Tkatchenko,~A. Combining Machine Learning and Computational Chemistry for Predictive Insights Into Chemical Systems. \emph{Chemical Reviews} \textbf{2021}, \emph{121}, 9816--9872, PMID: 34232033\relax
\mciteBstWouldAddEndPuncttrue
\mciteSetBstMidEndSepPunct{\mcitedefaultmidpunct}
{\mcitedefaultendpunct}{\mcitedefaultseppunct}\relax
\EndOfBibitem
\bibitem[OpenAI \latin{et~al.}(2024)OpenAI, Achiam, Adler, Agarwal, Ahmad, Akkaya, Aleman, Almeida, Altenschmidt, Altman, Anadkat, Avila, Babuschkin, Balaji, Balcom, Baltescu, Bao, Bavarian, Belgum, Bello, Berdine, Bernadett-Shapiro, Berner, Bogdonoff, Boiko, Boyd, Brakman, Brockman, Brooks, Brundage, Button, Cai, Campbell, Cann, Carey, Carlson, Carmichael, Chan, Chang, Chantzis, Chen, Chen, Chen, Chen, Chen, Chess, Cho, Chu, Chung, Cummings, Currier, Dai, Decareaux, Degry, Deutsch, Deville, Dhar, Dohan, Dowling, Dunning, Ecoffet, Eleti, Eloundou, Farhi, Fedus, Felix, Fishman, Forte, Fulford, Gao, Georges, Gibson, Goel, Gogineni, Goh, Gontijo-Lopes, Gordon, Grafstein, Gray, Greene, Gross, Gu, Guo, Hallacy, Han, Harris, He, Heaton, Heidecke, Hesse, Hickey, Hickey, Hoeschele, Houghton, Hsu, Hu, Hu, Huizinga, Jain, Jain, Jang, Jiang, Jiang, Jin, Jin, Jomoto, Jonn, Jun, Kaftan, Łukasz Kaiser, Kamali, Kanitscheider, Keskar, Khan, Kilpatrick, Kim, Kim, Kim, Kirchner, Kiros, Knight, Kokotajlo, Łukasz Kondraciuk,
  Kondrich, Konstantinidis, Kosic, Krueger, Kuo, Lampe, Lan, Lee, Leike, Leung, Levy, Li, Lim, Lin, Lin, Litwin, Lopez, Lowe, Lue, Makanju, Malfacini, Manning, Markov, Markovski, Martin, Mayer, Mayne, McGrew, McKinney, McLeavey, McMillan, McNeil, Medina, Mehta, Menick, Metz, Mishchenko, Mishkin, Monaco, Morikawa, Mossing, Mu, Murati, Murk, Mély, Nair, Nakano, Nayak, Neelakantan, Ngo, Noh, Ouyang, O'Keefe, Pachocki, Paino, Palermo, Pantuliano, Parascandolo, Parish, Parparita, Passos, Pavlov, Peng, Perelman, de~Avila Belbute~Peres, Petrov, de~Oliveira~Pinto, Michael, Pokorny, Pokrass, Pong, Powell, Power, Power, Proehl, Puri, Radford, Rae, Ramesh, Raymond, Real, Rimbach, Ross, Rotsted, Roussez, Ryder, Saltarelli, Sanders, Santurkar, Sastry, Schmidt, Schnurr, Schulman, Selsam, Sheppard, Sherbakov, Shieh, Shoker, Shyam, Sidor, Sigler, Simens, Sitkin, Slama, Sohl, Sokolowsky, Song, Staudacher, Such, Summers, Sutskever, Tang, Tezak, Thompson, Tillet, Tootoonchian, Tseng, Tuggle, Turley, Tworek, Uribe, Vallone,
  Vijayvergiya, Voss, Wainwright, Wang, Wang, Wang, Ward, Wei, Weinmann, Welihinda, Welinder, Weng, Weng, Wiethoff, Willner, Winter, Wolrich, Wong, Workman, Wu, Wu, Wu, Xiao, Xu, Yoo, Yu, Yuan, Zaremba, Zellers, Zhang, Zhang, Zhao, Zheng, Zhuang, Zhuk, and Zoph]{chatgpt4}
OpenAI \latin{et~al.}  GPT-4 Technical Report. 2024; \url{https://arxiv.org/abs/2303.08774}\relax
\mciteBstWouldAddEndPuncttrue
\mciteSetBstMidEndSepPunct{\mcitedefaultmidpunct}
{\mcitedefaultendpunct}{\mcitedefaultseppunct}\relax
\EndOfBibitem
\bibitem[Touvron \latin{et~al.}(2023)Touvron, Lavril, Izacard, Martinet, Lachaux, Lacroix, Rozière, Goyal, Hambro, Azhar, Rodriguez, Joulin, Grave, and Lample]{llama}
Touvron,~H.; Lavril,~T.; Izacard,~G.; Martinet,~X.; Lachaux,~M.-A.; Lacroix,~T.; Rozière,~B.; Goyal,~N.; Hambro,~E.; Azhar,~F.; Rodriguez,~A.; Joulin,~A.; Grave,~E.; Lample,~G. LLaMA: Open and Efficient Foundation Language Models. 2023; \url{https://arxiv.org/abs/2302.13971}\relax
\mciteBstWouldAddEndPuncttrue
\mciteSetBstMidEndSepPunct{\mcitedefaultmidpunct}
{\mcitedefaultendpunct}{\mcitedefaultseppunct}\relax
\EndOfBibitem
\bibitem[Wang \latin{et~al.}(2023)Wang, Xu, Li, and Barati~Farimani]{Wang2023}
Wang,~Y.; Xu,~C.; Li,~Z.; Barati~Farimani,~A. Denoise Pretraining on Nonequilibrium Molecules for Accurate and Transferable Neural Potentials. \emph{Journal of Chemical Theory and Computation} \textbf{2023}, \emph{19}, 5077--5087, PMID: 37390120\relax
\mciteBstWouldAddEndPuncttrue
\mciteSetBstMidEndSepPunct{\mcitedefaultmidpunct}
{\mcitedefaultendpunct}{\mcitedefaultseppunct}\relax
\EndOfBibitem
\bibitem[Xia \latin{et~al.}(2023)Xia, Zhao, Hu, Gao, Tan, Liu, Li, and Li]{Xia2023iclr}
Xia,~J.; Zhao,~C.; Hu,~B.; Gao,~Z.; Tan,~C.; Liu,~Y.; Li,~S.; Li,~S.~Z. Mole-{BERT}: Rethinking Pre-training Graph Neural Networks for Molecules. The Eleventh International Conference on Learning Representations. 2023\relax
\mciteBstWouldAddEndPuncttrue
\mciteSetBstMidEndSepPunct{\mcitedefaultmidpunct}
{\mcitedefaultendpunct}{\mcitedefaultseppunct}\relax
\EndOfBibitem
\bibitem[Xia \latin{et~al.}(2023)Xia, Zhu, Du, and Li]{Xia2023ijcai}
Xia,~J.; Zhu,~Y.; Du,~Y.; Li,~S.~Z. A Systematic Survey of Chemical Pre-trained Models. Proceedings of the Thirty-Second International Joint Conference on Artificial Intelligence, {IJCAI-23}. 2023; pp 6787--6795, Survey Track\relax
\mciteBstWouldAddEndPuncttrue
\mciteSetBstMidEndSepPunct{\mcitedefaultmidpunct}
{\mcitedefaultendpunct}{\mcitedefaultseppunct}\relax
\EndOfBibitem
\bibitem[Hu \latin{et~al.}(2020)Hu, Liu, Gomes, Zitnik, Liang, Pande, and Leskovec]{Hu2020Strategies}
Hu,~W.; Liu,~B.; Gomes,~J.; Zitnik,~M.; Liang,~P.; Pande,~V.; Leskovec,~J. Strategies for Pre-training Graph Neural Networks. International Conference on Learning Representations. 2020\relax
\mciteBstWouldAddEndPuncttrue
\mciteSetBstMidEndSepPunct{\mcitedefaultmidpunct}
{\mcitedefaultendpunct}{\mcitedefaultseppunct}\relax
\EndOfBibitem
\bibitem[Beck(2005)]{Beck2005}
Beck,~M.~E. Do Fukui Function Maxima Relate to Sites of Metabolism? A Critical Case Study. \emph{Journal of Chemical Information and Modeling} \textbf{2005}, \emph{45}, 273--282, PMID: 15807488\relax
\mciteBstWouldAddEndPuncttrue
\mciteSetBstMidEndSepPunct{\mcitedefaultmidpunct}
{\mcitedefaultendpunct}{\mcitedefaultseppunct}\relax
\EndOfBibitem
\bibitem[Wang \latin{et~al.}(2023)Wang, Wang, Wang, Ren, Chen, Li, Han, and Song]{quantumtox}
Wang,~X.; Wang,~L.; Wang,~S.; Ren,~Y.; Chen,~W.; Li,~X.; Han,~P.; Song,~T. QuantumTox: Utilizing quantum chemistry with ensemble learning for molecular toxicity prediction. \emph{Computers in Biology and Medicine} \textbf{2023}, \emph{157}, 106744\relax
\mciteBstWouldAddEndPuncttrue
\mciteSetBstMidEndSepPunct{\mcitedefaultmidpunct}
{\mcitedefaultendpunct}{\mcitedefaultseppunct}\relax
\EndOfBibitem
\bibitem[Beck and Schindler(2007)Beck, and Schindler]{fukuibook}
Beck,~M.~E.; Schindler,~M. \emph{Pesticide Chemistry}; John Wiley \& Sons, Ltd, 2007; Chapter 24, pp 227--238\relax
\mciteBstWouldAddEndPuncttrue
\mciteSetBstMidEndSepPunct{\mcitedefaultmidpunct}
{\mcitedefaultendpunct}{\mcitedefaultseppunct}\relax
\EndOfBibitem
\bibitem[Göller(2019)]{atomicdescriptordesign}
Göller,~A.~H. The art of atom descriptor design. \emph{Drug Discovery Today: Technologies} \textbf{2019}, \emph{32-33}, 37--43, Artificial Intelligence\relax
\mciteBstWouldAddEndPuncttrue
\mciteSetBstMidEndSepPunct{\mcitedefaultmidpunct}
{\mcitedefaultendpunct}{\mcitedefaultseppunct}\relax
\EndOfBibitem
\bibitem[Hoja \latin{et~al.}(2021)Hoja, Medrano~Sandonas, Ernst, Vazquez-Mayagoitia, DiStasio~Jr., and Tkatchenko]{Hoja2021}
Hoja,~J.; Medrano~Sandonas,~L.; Ernst,~B.~G.; Vazquez-Mayagoitia,~A.; DiStasio~Jr.,~R.~A.; Tkatchenko,~A. QM7-X, a comprehensive dataset of quantum-mechanical properties spanning the chemical space of small organic molecules. \emph{Scientific Data} \textbf{2021}, \emph{8}, 43\relax
\mciteBstWouldAddEndPuncttrue
\mciteSetBstMidEndSepPunct{\mcitedefaultmidpunct}
{\mcitedefaultendpunct}{\mcitedefaultseppunct}\relax
\EndOfBibitem
\bibitem[Medrano~Sandonas \latin{et~al.}(2024)Medrano~Sandonas, Van~Rompaey, Fallani, Hilfiker, Hahn, Perez-Benito, Verhoeven, Tresadern, Kurt~Wegner, Ceulemans, and Tkatchenko]{MedranoSandonas2024}
Medrano~Sandonas,~L.; Van~Rompaey,~D.; Fallani,~A.; Hilfiker,~M.; Hahn,~D.; Perez-Benito,~L.; Verhoeven,~J.; Tresadern,~G.; Kurt~Wegner,~J.; Ceulemans,~H.; Tkatchenko,~A. Dataset for quantum-mechanical exploration of conformers and solvent effects in large drug-like molecules. \emph{Scientific Data} \textbf{2024}, \emph{11}, 742\relax
\mciteBstWouldAddEndPuncttrue
\mciteSetBstMidEndSepPunct{\mcitedefaultmidpunct}
{\mcitedefaultendpunct}{\mcitedefaultseppunct}\relax
\EndOfBibitem
\bibitem[Isert \latin{et~al.}(2022)Isert, Atz, Jim{\'{e}}nez-Luna, and Schneider]{Isert2022}
Isert,~C.; Atz,~K.; Jim{\'{e}}nez-Luna,~J.; Schneider,~G. {QMugs, quantum mechanical properties of drug-like molecules}. \emph{Scientific Data} \textbf{2022}, \emph{9}\relax
\mciteBstWouldAddEndPuncttrue
\mciteSetBstMidEndSepPunct{\mcitedefaultmidpunct}
{\mcitedefaultendpunct}{\mcitedefaultseppunct}\relax
\EndOfBibitem
\bibitem[Kirklin \latin{et~al.}(2015)Kirklin, Saal, Meredig, Thompson, Doak, Aykol, R{\"u}hl, and Wolverton]{OQMD}
Kirklin,~S.; Saal,~J.~E.; Meredig,~B.; Thompson,~A.; Doak,~J.~W.; Aykol,~M.; R{\"u}hl,~S.; Wolverton,~C. The Open Quantum Materials Database (OQMD): assessing the accuracy of DFT formation energies. \emph{npj Computational Materials} \textbf{2015}, \emph{1}, 15010\relax
\mciteBstWouldAddEndPuncttrue
\mciteSetBstMidEndSepPunct{\mcitedefaultmidpunct}
{\mcitedefaultendpunct}{\mcitedefaultseppunct}\relax
\EndOfBibitem
\bibitem[Nakata \latin{et~al.}(2020)Nakata, Shimazaki, Hashimoto, and Maeda]{NakataPM6}
Nakata,~M.; Shimazaki,~T.; Hashimoto,~M.; Maeda,~T. PubChemQC PM6: Data Sets of 221 Million Molecules with Optimized Molecular Geometries and Electronic Properties. \emph{Journal of Chemical Information and Modeling} \textbf{2020}, \emph{60}, 5891--5899, PMID: 33104339\relax
\mciteBstWouldAddEndPuncttrue
\mciteSetBstMidEndSepPunct{\mcitedefaultmidpunct}
{\mcitedefaultendpunct}{\mcitedefaultseppunct}\relax
\EndOfBibitem
\bibitem[Chanussot \latin{et~al.}(2021)Chanussot, Das, Goyal, Lavril, Shuaibi, Riviere, Tran, Heras-Domingo, Ho, Hu, Palizhati, Sriram, Wood, Yoon, Parikh, Zitnick, and Ulissi]{OpenCatalyst}
Chanussot,~L. \latin{et~al.}  Open Catalyst 2020 (OC20) Dataset and Community Challenges. \emph{ACS Catalysis} \textbf{2021}, \emph{11}, 6059--6072\relax
\mciteBstWouldAddEndPuncttrue
\mciteSetBstMidEndSepPunct{\mcitedefaultmidpunct}
{\mcitedefaultendpunct}{\mcitedefaultseppunct}\relax
\EndOfBibitem
\bibitem[Chmiela \latin{et~al.}(2017)Chmiela, Tkatchenko, Sauceda, Poltavsky, Schütt, and Müller]{sgdml}
Chmiela,~S.; Tkatchenko,~A.; Sauceda,~H.~E.; Poltavsky,~I.; Schütt,~K.~T.; Müller,~K.-R. Machine learning of accurate energy-conserving molecular force fields. \emph{Science Advances} \textbf{2017}, \emph{3}, e1603015\relax
\mciteBstWouldAddEndPuncttrue
\mciteSetBstMidEndSepPunct{\mcitedefaultmidpunct}
{\mcitedefaultendpunct}{\mcitedefaultseppunct}\relax
\EndOfBibitem
\bibitem[Hachmann \latin{et~al.}(2011)Hachmann, Olivares-Amaya, Atahan-Evrenk, Amador-Bedolla, Sánchez-Carrera, Gold-Parker, Vogt, Brockway, and Aspuru-Guzik]{HarvardCleanEnergy}
Hachmann,~J.; Olivares-Amaya,~R.; Atahan-Evrenk,~S.; Amador-Bedolla,~C.; Sánchez-Carrera,~R.~S.; Gold-Parker,~A.; Vogt,~L.; Brockway,~A.~M.; Aspuru-Guzik,~A. The Harvard Clean Energy Project: Large-Scale Computational Screening and Design of Organic Photovoltaics on the World Community Grid. \emph{The Journal of Physical Chemistry Letters} \textbf{2011}, \emph{2}, 2241--2251\relax
\mciteBstWouldAddEndPuncttrue
\mciteSetBstMidEndSepPunct{\mcitedefaultmidpunct}
{\mcitedefaultendpunct}{\mcitedefaultseppunct}\relax
\EndOfBibitem
\bibitem[Mobley and Guthrie(2014)Mobley, and Guthrie]{FreeSolv}
Mobley,~D.~L.; Guthrie,~J.~P. FreeSolv: a database of experimental and calculated hydration free energies, with input files. \emph{Journal of Computer-Aided Molecular Design} \textbf{2014}, \emph{28}, 711--720\relax
\mciteBstWouldAddEndPuncttrue
\mciteSetBstMidEndSepPunct{\mcitedefaultmidpunct}
{\mcitedefaultendpunct}{\mcitedefaultseppunct}\relax
\EndOfBibitem
\bibitem[Gilson \latin{et~al.}(2015)Gilson, Liu, Baitaluk, Nicola, Hwang, and Chong]{BindingDB}
Gilson,~M.~K.; Liu,~T.; Baitaluk,~M.; Nicola,~G.; Hwang,~L.; Chong,~J. {BindingDB in 2015: A public database for medicinal chemistry, computational chemistry and systems pharmacology}. \emph{Nucleic Acids Research} \textbf{2015}, \emph{44}, D1045--D1053\relax
\mciteBstWouldAddEndPuncttrue
\mciteSetBstMidEndSepPunct{\mcitedefaultmidpunct}
{\mcitedefaultendpunct}{\mcitedefaultseppunct}\relax
\EndOfBibitem
\bibitem[Smith \latin{et~al.}(2018)Smith, Nebgen, Lubbers, Isayev, and Roitberg]{Ani1x}
Smith,~J.~S.; Nebgen,~B.; Lubbers,~N.; Isayev,~O.; Roitberg,~A.~E. {Less is more: Sampling chemical space with active learning}. \emph{The Journal of Chemical Physics} \textbf{2018}, \emph{148}, 241733\relax
\mciteBstWouldAddEndPuncttrue
\mciteSetBstMidEndSepPunct{\mcitedefaultmidpunct}
{\mcitedefaultendpunct}{\mcitedefaultseppunct}\relax
\EndOfBibitem
\bibitem[Smith \latin{et~al.}(2017)Smith, Isayev, and Roitberg]{Ani1}
Smith,~J.~S.; Isayev,~O.; Roitberg,~A.~E. ANI-1, A data set of 20 million calculated off-equilibrium conformations for organic molecules. \emph{Scientific Data} \textbf{2017}, \emph{4}, 170193\relax
\mciteBstWouldAddEndPuncttrue
\mciteSetBstMidEndSepPunct{\mcitedefaultmidpunct}
{\mcitedefaultendpunct}{\mcitedefaultseppunct}\relax
\EndOfBibitem
\bibitem[Kläser \latin{et~al.}(2024)Kläser, Banaszewski, Maddrell-Mander, McLean, Müller, Parviz, Huang, and Fitzgibbon]{minimol}
Kläser,~K.; Banaszewski,~B.; Maddrell-Mander,~S.; McLean,~C.; Müller,~L.; Parviz,~A.; Huang,~S.; Fitzgibbon,~A. $\texttt{MiniMol}$: A Parameter-Efficient Foundation Model for Molecular Learning. 2024; \url{https://arxiv.org/abs/2404.14986}\relax
\mciteBstWouldAddEndPuncttrue
\mciteSetBstMidEndSepPunct{\mcitedefaultmidpunct}
{\mcitedefaultendpunct}{\mcitedefaultseppunct}\relax
\EndOfBibitem
\bibitem[Kim \latin{et~al.}(2024)Kim, Chang, Ji, and Joung]{Kim2024}
Kim,~J.; Chang,~W.; Ji,~H.; Joung,~I. Quantum-Informed Molecular Representation Learning Enhancing ADMET Property Prediction. \emph{Journal of Chemical Information and Modeling} \textbf{2024}, \emph{64}, 5028--5040, PMID: 38916580\relax
\mciteBstWouldAddEndPuncttrue
\mciteSetBstMidEndSepPunct{\mcitedefaultmidpunct}
{\mcitedefaultendpunct}{\mcitedefaultseppunct}\relax
\EndOfBibitem
\bibitem[Raja \latin{et~al.}(2024)Raja, Zhao, Tyrchan, Nittinger, Bronstein, Deane, and Morris]{raja2024on}
Raja,~A.; Zhao,~H.; Tyrchan,~C.; Nittinger,~E.; Bronstein,~M.~M.; Deane,~C.; Morris,~G.~M. On the Effectiveness of Quantum Chemistry Pre-training for Pharmacological Property Prediction. ICML 2024 AI for Science Workshop. 2024\relax
\mciteBstWouldAddEndPuncttrue
\mciteSetBstMidEndSepPunct{\mcitedefaultmidpunct}
{\mcitedefaultendpunct}{\mcitedefaultseppunct}\relax
\EndOfBibitem
\bibitem[Ying \latin{et~al.}(2021)Ying, Cai, Luo, Zheng, Ke, He, Shen, and Liu]{Ying:21}
Ying,~C.; Cai,~T.; Luo,~S.; Zheng,~S.; Ke,~G.; He,~D.; Shen,~Y.; Liu,~T.-Y. Do Transformers Really Perform Badly for Graph Representation? Advances in Neural Information Processing Systems. 2021; pp 28877--28888\relax
\mciteBstWouldAddEndPuncttrue
\mciteSetBstMidEndSepPunct{\mcitedefaultmidpunct}
{\mcitedefaultendpunct}{\mcitedefaultseppunct}\relax
\EndOfBibitem
\bibitem[Nugmanov \latin{et~al.}(2022)Nugmanov, Dyubankova, Gedich, and Wegner]{Nugmanov2022}
Nugmanov,~R.; Dyubankova,~N.; Gedich,~A.; Wegner,~J.~K. Bidirectional Graphormer for Reactivity Understanding: Neural Network Trained to Reaction Atom-to-Atom Mapping Task. \emph{Journal of Chemical Information and Modeling} \textbf{2022}, \emph{62}, 3307--3315, PMID: 35792579\relax
\mciteBstWouldAddEndPuncttrue
\mciteSetBstMidEndSepPunct{\mcitedefaultmidpunct}
{\mcitedefaultendpunct}{\mcitedefaultseppunct}\relax
\EndOfBibitem
\bibitem[M{\"u}ller \latin{et~al.}(2024)M{\"u}ller, Galkin, Morris, and Ramp{\'a}{\v{s}}ek]{muller2024attending}
M{\"u}ller,~L.; Galkin,~M.; Morris,~C.; Ramp{\'a}{\v{s}}ek,~L. Attending to Graph Transformers. \emph{Transactions on Machine Learning Research} \textbf{2024}, \relax
\mciteBstWouldAddEndPunctfalse
\mciteSetBstMidEndSepPunct{\mcitedefaultmidpunct}
{}{\mcitedefaultseppunct}\relax
\EndOfBibitem
\bibitem[Guan \latin{et~al.}(2021)Guan, Coley, Wu, Ranasinghe, Heid, Struble, Pattanaik, Green, and Jensen]{Guan2021}
Guan,~Y.; Coley,~C.~W.; Wu,~H.; Ranasinghe,~D.; Heid,~E.; Struble,~T.~J.; Pattanaik,~L.; Green,~W.~H.; Jensen,~K.~F. {Regio-selectivity prediction with a machine-learned reaction representation and on-the-fly quantum mechanical descriptors}. \emph{Chemical Science} \textbf{2021}, \emph{12}, 2198--2208\relax
\mciteBstWouldAddEndPuncttrue
\mciteSetBstMidEndSepPunct{\mcitedefaultmidpunct}
{\mcitedefaultendpunct}{\mcitedefaultseppunct}\relax
\EndOfBibitem
\bibitem[Nakata and Shimazaki(2017)Nakata, and Shimazaki]{Nakata2017}
Nakata,~M.; Shimazaki,~T. {PubChemQC Project: A Large-Scale First-Principles Electronic Structure Database for Data-Driven Chemistry}. \emph{Journal of Chemical Information and Modeling} \textbf{2017}, \emph{57}, 1300--1308\relax
\mciteBstWouldAddEndPuncttrue
\mciteSetBstMidEndSepPunct{\mcitedefaultmidpunct}
{\mcitedefaultendpunct}{\mcitedefaultseppunct}\relax
\EndOfBibitem
\bibitem[Huang \latin{et~al.}(2022)Huang, Fu, Gao, Zhao, Roohani, Leskovec, Coley, Xiao, Sun, and Zitnik]{Huang2022}
Huang,~K.; Fu,~T.; Gao,~W.; Zhao,~Y.; Roohani,~Y.; Leskovec,~J.; Coley,~C.~W.; Xiao,~C.; Sun,~J.; Zitnik,~M. {Artificial intelligence foundation for therapeutic science}. \emph{Nature Chemical Biology} \textbf{2022}, \emph{18}, 1033--1036\relax
\mciteBstWouldAddEndPuncttrue
\mciteSetBstMidEndSepPunct{\mcitedefaultmidpunct}
{\mcitedefaultendpunct}{\mcitedefaultseppunct}\relax
\EndOfBibitem
\bibitem[Abnar and Zuidema(2020)Abnar, and Zuidema]{rollout}
Abnar,~S.; Zuidema,~W.~H. Quantifying Attention Flow in Transformers. 2020; \url{https://arxiv.org/abs/2005.00928}\relax
\mciteBstWouldAddEndPuncttrue
\mciteSetBstMidEndSepPunct{\mcitedefaultmidpunct}
{\mcitedefaultendpunct}{\mcitedefaultseppunct}\relax
\EndOfBibitem
\bibitem[Fabian \latin{et~al.}(2020)Fabian, Edlich, Gaspar, Segler, Meyers, Fiscato, and Ahmed]{Fabian2020}
Fabian,~B.; Edlich,~T.; Gaspar,~H.; Segler,~M. H.~S.; Meyers,~J.; Fiscato,~M.; Ahmed,~M. Molecular representation learning with language models and domain-relevant auxiliary tasks. \emph{CoRR} \textbf{2020}, \emph{abs/2011.13230}\relax
\mciteBstWouldAddEndPuncttrue
\mciteSetBstMidEndSepPunct{\mcitedefaultmidpunct}
{\mcitedefaultendpunct}{\mcitedefaultseppunct}\relax
\EndOfBibitem
\bibitem[Dong \latin{et~al.}(2023)Dong, Cordonnier, and Loukas]{attentionrank}
Dong,~Y.; Cordonnier,~J.-B.; Loukas,~A. Attention is Not All You Need: Pure Attention Loses Rank Doubly Exponentially with Depth. 2023; \url{https://arxiv.org/abs/2103.03404}\relax
\mciteBstWouldAddEndPuncttrue
\mciteSetBstMidEndSepPunct{\mcitedefaultmidpunct}
{\mcitedefaultendpunct}{\mcitedefaultseppunct}\relax
\EndOfBibitem
\bibitem[Topping \latin{et~al.}(2022)Topping, Giovanni, Chamberlain, Dong, and Bronstein]{oversquashing_bronstein}
Topping,~J.; Giovanni,~F.~D.; Chamberlain,~B.~P.; Dong,~X.; Bronstein,~M.~M. Understanding over-squashing and bottlenecks on graphs via curvature. International Conference on Learning Representations. 2022\relax
\mciteBstWouldAddEndPuncttrue
\mciteSetBstMidEndSepPunct{\mcitedefaultmidpunct}
{\mcitedefaultendpunct}{\mcitedefaultseppunct}\relax
\EndOfBibitem
\bibitem[Arjona-Medina and Nugmanov(2024)Arjona-Medina, and Nugmanov]{joseramil}
Arjona-Medina,~J.; Nugmanov,~R. Analysis of Atom-level pretraining with Quantum Mechanics (QM) data for Graph Neural Networks Molecular property models. 2024; \url{https://arxiv.org/abs/2405.14837}\relax
\mciteBstWouldAddEndPuncttrue
\mciteSetBstMidEndSepPunct{\mcitedefaultmidpunct}
{\mcitedefaultendpunct}{\mcitedefaultseppunct}\relax
\EndOfBibitem
\bibitem[Fallani \latin{et~al.}(2025)Fallani, Arjona-Medina, Chernichenko, Nugmanov, Wegner, and Tkatchenko]{icann}
Fallani,~A.; Arjona-Medina,~J.; Chernichenko,~K.; Nugmanov,~R.; Wegner,~J.~K.; Tkatchenko,~A. Atom-Level Quantum Pretraining Enhances the Spectral Perception of Molecular Graphs in Graphormer. AI in Drug Discovery. Cham, 2025; pp 71--81\relax
\mciteBstWouldAddEndPuncttrue
\mciteSetBstMidEndSepPunct{\mcitedefaultmidpunct}
{\mcitedefaultendpunct}{\mcitedefaultseppunct}\relax
\EndOfBibitem
\bibitem[Fabian \latin{et~al.}(2020)Fabian, Edlich, Gaspar, Segler, Meyers, Fiscato, and Ahmed]{Fabian:20}
Fabian,~B.; Edlich,~T.; Gaspar,~H.; Segler,~M.; Meyers,~J.; Fiscato,~M.; Ahmed,~M. Molecular representation learning with language models and domain-relevant auxiliary tasks. Proc. NeurIPS 2020 Workshop on Machine Learning for Molecules. 2020\relax
\mciteBstWouldAddEndPuncttrue
\mciteSetBstMidEndSepPunct{\mcitedefaultmidpunct}
{\mcitedefaultendpunct}{\mcitedefaultseppunct}\relax
\EndOfBibitem
\bibitem[Devlin \latin{et~al.}(2018)Devlin, Chang, Lee, and Toutanova]{BERT}
Devlin,~J.; Chang,~M.-W.; Lee,~K.; Toutanova,~K. {BERT: Pre-training of Deep Bidirectional Transformers for Language Understanding}. \textbf{2018}, \relax
\mciteBstWouldAddEndPunctfalse
\mciteSetBstMidEndSepPunct{\mcitedefaultmidpunct}
{}{\mcitedefaultseppunct}\relax
\EndOfBibitem
\bibitem[Zou and Hastie(2005)Zou, and Hastie]{elasticnet}
Zou,~H.; Hastie,~T. {Regularization and Variable Selection Via the Elastic Net}. \emph{Journal of the Royal Statistical Society Series B: Statistical Methodology} \textbf{2005}, \emph{67}, 301--320\relax
\mciteBstWouldAddEndPuncttrue
\mciteSetBstMidEndSepPunct{\mcitedefaultmidpunct}
{\mcitedefaultendpunct}{\mcitedefaultseppunct}\relax
\EndOfBibitem
\bibitem[{Van Damme} and Bultinck(2009){Van Damme}, and Bultinck]{VanDamme2009}
{Van Damme},~S.; Bultinck,~P. {Conceptual DFT properties-based 3D QSAR: Analysis of inhibitors of the nicotine metabolizing CYP2A6 enzyme}. \emph{Journal of Computational Chemistry} \textbf{2009}, \emph{30}, 1749--1757\relax
\mciteBstWouldAddEndPuncttrue
\mciteSetBstMidEndSepPunct{\mcitedefaultmidpunct}
{\mcitedefaultendpunct}{\mcitedefaultseppunct}\relax
\EndOfBibitem
\bibitem[Lewis(1997)]{Lewis1997}
Lewis,~D.~F. {Quantitative structure-activity relationships in substrates, inducers, and inhibitors of cytochrome P4501 (CYP1)}. \emph{Drug Metabolism Reviews} \textbf{1997}, \emph{29}, 589--650\relax
\mciteBstWouldAddEndPuncttrue
\mciteSetBstMidEndSepPunct{\mcitedefaultmidpunct}
{\mcitedefaultendpunct}{\mcitedefaultseppunct}\relax
\EndOfBibitem
\bibitem[Ai \latin{et~al.}(2010)Ai, Li, Wang, Li, Dong, Ge, and Yang]{Ai2010}
Ai,~C.; Li,~Y.; Wang,~Y.; Li,~W.; Dong,~P.; Ge,~G.; Yang,~L. {Investigation of binding features: Effects on the interaction between CYP2A6 and inhibitors}. \emph{Journal of Computational Chemistry} \textbf{2010}, \emph{31}, 1822--1831\relax
\mciteBstWouldAddEndPuncttrue
\mciteSetBstMidEndSepPunct{\mcitedefaultmidpunct}
{\mcitedefaultendpunct}{\mcitedefaultseppunct}\relax
\EndOfBibitem
\bibitem[Kaya and Kaya(2015)Kaya, and Kaya]{Kaya2015}
Kaya,~S.; Kaya,~C. {A new method for calculation of molecular hardness: A theoretical study}. \emph{Computational and Theoretical Chemistry} \textbf{2015}, \emph{1060}, 66--70\relax
\mciteBstWouldAddEndPuncttrue
\mciteSetBstMidEndSepPunct{\mcitedefaultmidpunct}
{\mcitedefaultendpunct}{\mcitedefaultseppunct}\relax
\EndOfBibitem
\bibitem[Bruna \latin{et~al.}(2014)Bruna, Zaremba, Szlam, and LeCun]{lecunnspectral}
Bruna,~J.; Zaremba,~W.; Szlam,~A.; LeCun,~Y. Spectral Networks and Locally Connected Networks on Graphs. 2014; \url{https://arxiv.org/abs/1312.6203}\relax
\mciteBstWouldAddEndPuncttrue
\mciteSetBstMidEndSepPunct{\mcitedefaultmidpunct}
{\mcitedefaultendpunct}{\mcitedefaultseppunct}\relax
\EndOfBibitem
\bibitem[Kipf and Welling(2017)Kipf, and Welling]{wellingspectral}
Kipf,~T.~N.; Welling,~M. Semi-Supervised Classification with Graph Convolutional Networks. 2017; \url{https://arxiv.org/abs/1609.02907}\relax
\mciteBstWouldAddEndPuncttrue
\mciteSetBstMidEndSepPunct{\mcitedefaultmidpunct}
{\mcitedefaultendpunct}{\mcitedefaultseppunct}\relax
\EndOfBibitem
\bibitem[Hammond \latin{et~al.}(2009)Hammond, Vandergheynst, and Gribonval]{waveletspectral}
Hammond,~D.~K.; Vandergheynst,~P.; Gribonval,~R. Wavelets on Graphs via Spectral Graph Theory. 2009; \url{https://arxiv.org/abs/0912.3848}\relax
\mciteBstWouldAddEndPuncttrue
\mciteSetBstMidEndSepPunct{\mcitedefaultmidpunct}
{\mcitedefaultendpunct}{\mcitedefaultseppunct}\relax
\EndOfBibitem
\bibitem[Defferrard \latin{et~al.}(2017)Defferrard, Bresson, and Vandergheynst]{fastspectral}
Defferrard,~M.; Bresson,~X.; Vandergheynst,~P. Convolutional Neural Networks on Graphs with Fast Localized Spectral Filtering. 2017; \url{https://arxiv.org/abs/1606.09375}\relax
\mciteBstWouldAddEndPuncttrue
\mciteSetBstMidEndSepPunct{\mcitedefaultmidpunct}
{\mcitedefaultendpunct}{\mcitedefaultseppunct}\relax
\EndOfBibitem
\end{mcitethebibliography}

\end{document}




\section{Motivation for rollout as model proxy}\label{motivation_rollout}
If we want to do a spectral analysis of the considered network (i.e. which combination of input components are extracted), the choice of Attention Rollout is a rather natural one. This matrix, in fact, is a good proxy to understand how the information coming from the input is mixed throughout the network. 
One can retrieve this intuition by considering the action of a Self Attention Network (SAN) architecture with skip connections at layer $l$ on the input $X_{l-1}\in \mathbb{R}^{n\times d}$ as formalized in \cite{transf_oversmoothing}. This update can be written as:
\begin{equation}
X_l = X_{l-1} + A_{l} X_{l-1} H^T,
\end{equation}
where $A_{l}\in\mathbb{R}^{n\times n}$ is the input-dependent attention matrix at layer $l$ (hence dependent on $X_{l-1}$) and $H\in \mathbb{R}^{d\times d}$ is the combination of the value matrix $W_V$ and the projection head $W_{proj}$ (namely $H = W_{proj}^TW_V^T$) and does not depend on the input $X_l$ for any $l$. If one is only interested in how the input information gets mixed between tokens, one can notice that the action of $H$ on $X_l$ is a token-wise transformation and study the case for $H=\mathbb{I}$ (notice that for the same reason we did not consider the activation function) to obtain:
\begin{equation} \label{eq1}
\begin{split}
X_L & = (\mathbb{I}+ A_{l})X_{l-1}  \\
 & = \left[\prod_{l=L}^{l=1}(\mathbb{I}+ A_{l})\right]X \\
 & =2^L \Tilde{A}_L X
\end{split}
\end{equation}
where $\Tilde{A}_L = \prod_{l=L}^{l=1}(\mathbb{I}+ A_{l})$ is the attention rollout matrix as defined in \cite{rollout}. Notice that this observation is valid only for a single-headed attention case. To handle multi-headed networks here we decide to follow the definition of $\Tilde{A}_L$ from the previous section and just consider the corresponding single-headed case for which  $A_{l}\rightarrow\langle A_{l}\rangle_{heads}$. Considering $\Tilde{A}$ as a proxy for the model is of course a strong simplification of the network but as we will see it manages to be useful to gain spectral insights.

\section{Spectral properties}\label{rollout_spectral_properties}
Now that we motivated the use of the Attention Rollout matrix as a proxy for the transformer network, we look at its eigenvalues and eigenvectors in order to understand along which directions the input $X$ is decomposed. 
For simplicity we will make use of the bra-ket notation.
Given the Rollout matrix $\Tilde{A}$ (we drop the subscript relative to network depth for simplicity of notation) we can write:
\begin{equation}
\Tilde{A} = \sum_{i=0}^{N-1} a_i\ket{a_i}\bra{a_i} 
\end{equation}
where $a_i$ are the eigenvalues of $\Tilde{A}$ and where$\ket{a_i}\bra{a_i}$ are the projectors on the associated eigenvectors $\ket{a_i}$.
It is important to state here that $\Tilde{A}$ are in general non-symmetric positive definite matrices, hence their eigenvectors can be non-orthogonal and their eigenvalues are complex numbers with positive real parts.
As additional observation, considering in the definition of $\Tilde{A}$ that attention matrices are row-stochastic matrices, we can say that $\Tilde{A}$ is also row-stochastic. This fact implies that $\Tilde{A} \ket{\textbf{1}} = \ket{\textbf{1}}$ (with $\ket{\textbf{1}}$ normalized vector with all equal components) and that $|a_i|\leq1$ $\forall $ $i$, hence if we sort the eigenvalues in descending order of magnitude we can say that $a_0 = 1$ and that $\ket{a_0} = \ket{\textbf{1}}$. If we consider the spectrum of the molecular graph Laplacian $L$ as:
\begin{equation}
    L = \sum_{i = 0}^{N-1} l_i\ket{l_i}\bra{l_i}
\end{equation}
with $l_0\leq l_1\leq ...\leq l_{N-1}$, we see that because of basic properties of the graph Laplacian $L$ \cite{Fiedler1973} we have $\braket{a_0|l_0}=1$.

\section{Filtered convolution-like behaviour and connection with graph signal oversmoothing}\label{spectral_overlap_conv}
More in detail, one can analyse the matrix of the normalized eigenvectors overlap $C_{ij} = \braket{a_i|l_j}$, and consider the spectrum $\{a_i\}$ as the weight associated to this decomposition. If we now denote with $\mathcal{U}$ the set of $i$ for which $\max_{j}C_{ij}>0.9$ (threshold for overlap close to 1), then provided $\eta = \frac{\sum_{i\in\mathcal{U}\setminus 0} |a_i|}{\sum_{i=1}^{i=N-1} |a_i|}\sim 1$ we can write the action of the network proxy $\Tilde{A}$ on the input $\ket{x}$ as:
\begin{equation}
\Tilde{A}\ket{x} \sim \sum_{i \in \mathcal{U}} a_i \ket{l_i}\braket{l_i|x},
\label{convolution}
\end{equation}
which is interpretable as a filtered graph convolution where the relevant rollout eigenvalues $\{a_i | i\in\mathcal{U}\}$ play the role of filter. Notice that if this approximation is valid ($\eta\sim 1$), we can see the number of Laplacian eigenmodes considered in this sum as the \textit{bandwidth} of a graph convolutional filter. Notice that independently of wether this approximation is valid or not (that is to say if there is some part of the spectrum that is not related to Laplacian eigenmodes) the set of $\{a_i|i\in\mathcal{U}\}$ can be associated to the amount of graph information captured by the model, effectively meaning that $\zeta$ can also be seen as measure of oversmoothing\cite{survey_oversmoothing, transf_oversmoothing} (or lack thereof) in the sense of graph-spectral information for these models.
In order to study this effect on Graphormer we need some tailoring. In this network, in fact, we have a class token used for molecule level modeling which is not present in the molecular graph. To handle this problesm, we consider first the full attention to obtain the rollout matrix $\Tilde{A}$ and preserve row-stochasticity properties, but then when considering the eigenvectors $\ket{a_i}$ we slice only the components associated with atoms and then normalize before taking inner products with Laplacian eigenvectors. This can be seen in both models as looking at the token subspace relative to the molecular graph nodes.

\section{Latent expressivity for single atomic property pretrainings}
\begin{figure}
    \centering
    \includegraphics[width=\textwidth]{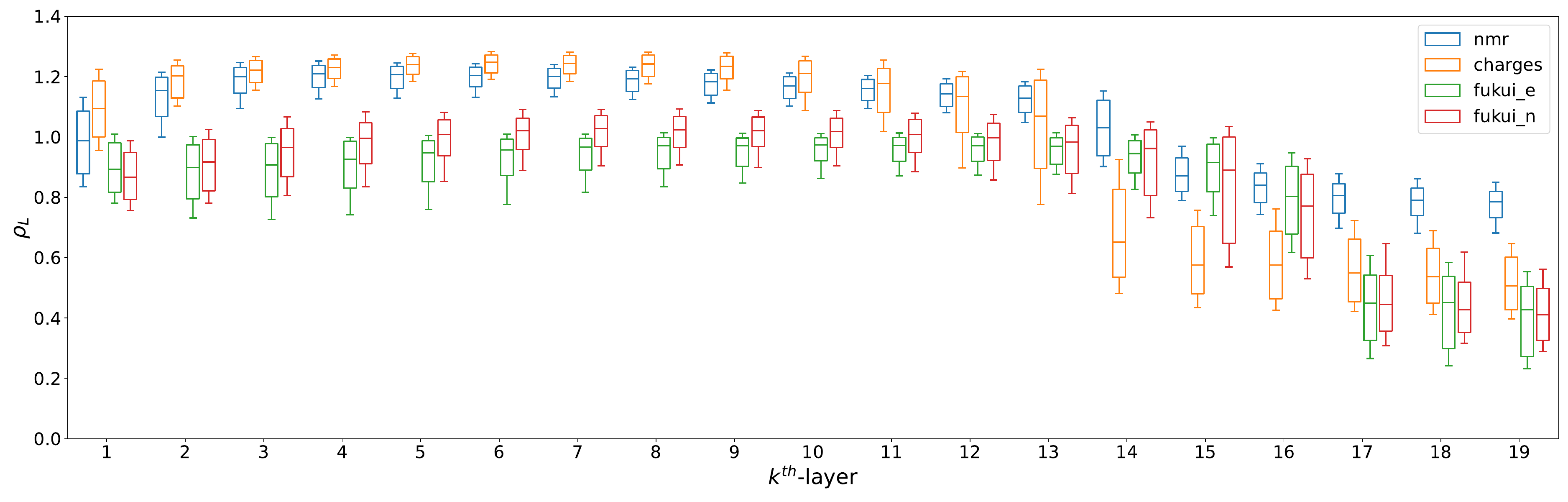}
    \caption{Boxplots of $\rho_L$ computed at each layer except the last one for a sample of 2200 structures extracted uniformly from all the fine-tuning test sets (100 structures for each of the 22 tasks). We report this quantity for the models pretrained on each individual atomic property. The whiskers go from the 15th percentile to the 85th for better visualization of trends and outliers are excluded for the same reason.}
    \label{fig:quantum_rank}
\end{figure}

In Fig. \ref{fig:quantum_rank} we we report the analysis of $\rho_L$ as defined in the main text, for all the models pretrained on each single atomic quantum property. We can see how models pretrained on nmr and charges have a similar trend across layers as model pretrained on all atomic properties, while models pretrained on fukui functions, while providing a similar trend, achieve a lower maximum value in the initial part of the network.

\section{Implementation Details}
The Graphormer encoder network employed in this work is composed of $20$ self-attention layers, with embedding dimension of $256$ and $32$ heads and dimension of the feedforward component in each layer of $512$. Transformer pre-norm is employed together with a dropout of $0.1$. The optimizer used is AdamW with the default value of weight decay set to $0.01$, and we employed mixed-precision training for  both pretraining and fine-tuning. For what concerns the pretraining phase no test set was used in favour of a .8/.2 training/validation split. We employed L1 loss and a constant learning rate of $10^-4$ with a patience of $100$ epochs for early stopping across all cases. For atomic QM pretraining we employed a batch size of $100$ while given the higher number of samples and the duration of training we employed a batch size of $1000$ for the pretraining on HOMO-LUMO gap. No scaling of the labels is employed at this stage with the exception of NMR shielding constants where we divided by a constant factor of $100$ as otherwise convergence would have been too slow. For the fine-tuning cases the hyperparameters used in each downstream task are the same: the batch size used is $32$, while for what concerns the learning rate a triangular cyclic scheduling was employed with a minimum value of $2\times 10^{-5}$ and a maximum value of $2\times 10^{-4}$. The training is stopped with an early stopping criterion with patience of $200$. The loss used for regression tasks is L1 loss, while for classification tasks a censored regression approach is taken using again L1 loss with right censor set at $0$ for negative examples and left censor set at $1$ for positive examples. For what concerns regression labels, given the diversity of the tasks we opted for a standard scaling. We conclude this section providing the dataset dimension for each individual downstream task dataset in table \ref{tab:dataset-sizes}.

\begin{table}
    \centering
    \caption{Dataset sizes for different tasks}
    \label{tab:dataset-sizes}
    \begin{tabular}{ll}
        \toprule
        Task & Dataset Size \\
        \midrule
        caco2\_wang & 910 \\
        hia\_hou & 578 \\
        pgp\_broccatelli & 1218 \\
        bioavailability\_ma & 640 \\
        lipophilicity\_astrazeneca & 4200 \\
        solubility\_aqsoldb & 9982 \\
        bbb\_martins & 2030 \\
        ppbr\_az & 2790 \\
        vdss\_lombardo & 1130 \\
        cyp2d6\_veith & 13130 \\
        cyp3a4\_veith & 12328 \\
        cyp2c9\_veith & 12092 \\
        cyp2d6\_substrate\_carbonmangels & 667 \\
        cyp3a4\_substrate\_carbonmangels & 670 \\
        cyp2c9\_substrate\_carbonmangels & 669 \\
        half\_life\_obach & 667 \\
        clearance\_microsome\_az & 1102 \\
        clearance\_hepatocyte\_az & 1213 \\
        herg & 655 \\
        ames & 7278 \\
        dili & 475 \\
        ld50\_zhu & 7385 \\
        \bottomrule
    \end{tabular}
\end{table}

\section{Regression modelling of JNJ data for human liver microsome intrinsic clearance}
The dataset contains 138439 unique compounds.
Measurements are reported for two assays: for 54611 compounds only for assay 1, for 66631 only for assay 2, and for 23358 compounds for both assays. For the multitask regression we used L2 loss masking out the missing values for the cases reporting only one clearance value.

\bibliography{bibliografia}